\documentclass[aip,cha,amsmath,amssymb,reprint,floatfix]{revtex4-2}
\usepackage[T1]{fontenc}
\usepackage[utf8]{inputenc}
\usepackage{graphicx,bm,amsthm,booktabs,microtype,capt-of}
\usepackage[colorlinks=true,citecolor=blue,linkcolor=blue,urlcolor=blue]{hyperref}
\graphicspath{{figures_final/}{./}}
\newcommand{\manuscriptfigure}[2][]{%
 \IfFileExists{figures_final/#2}{\includegraphics[#1]{figures_final/#2}}{%
  \IfFileExists{#2}{\includegraphics[#1]{#2}}{%
   \fbox{\parbox[c][.12\textheight][c]{.88\linewidth}{\centering\small
    Figure file: \texttt{\detokenize{#2}}\\[4pt]
    The complete figure appears in the accompanying PDF and source bundle.}}}}}

\hypersetup{pdftitle={Inference for stochastic differential equations driven by weighted sub-fractional Brownian motion using neural networks and the Euler approximation},pdfauthor={J. H. Ramirez-Gonzalez}}
\newtheorem{proposition}{Proposition}

\begin{document}
\raggedbottom
\title[Neural inference for SDEs driven by wsBm]{Inference for stochastic differential equations driven by weighted sub-fractional Brownian motion using neural networks and the Euler approximation}
\author{J.~H.~Ram\'irez-Gonz\'alez}
\email{hermenegildo.ramirez@usp.br}
\affiliation{Institute of Mathematics and Statistics, University of S\~ao Paulo (IME--USP), S\~ao Paulo, Brazil}
\date{September 28, 2026}
\begin{abstract}
We consider the estimation of drift, diffusion, and noise covariance from discrete observations of stochastic differential equations driven by Gaussian processes. For a fixed observation horizon $T>0$ and a known initial state $x_0\in\mathbb R$, we study
\begin{equation*}
 dX_t=a(X_t)\,dt+\sigma(X_t)\,dZ_t^{\beta,f},
 \qquad X_0=x_0,\quad 0\leq t\leq T.
\end{equation*}
\smallskip\noindent
Here $a:\mathbb R\to\mathbb R$ is the drift coefficient, $\sigma:\mathbb R\to(0,\infty)$ is the diffusion coefficient, and $Z^{\beta,f}$ is a centered Gaussian process from the weighted sub-fractional Brownian family, with covariance
\begin{equation*}
 \operatorname{Cov}(Z_s^{\beta,f},Z_t^{\beta,f})
 =\int_0^{s\wedge t} f(r)q_\beta(s-r,t-r)\,dr,
 \qquad 0\leq s,t\leq T.
\end{equation*}
\smallskip\noindent
Here $s\wedge t=\min\{s,t\}$. The temporal weight $f:[0,T]\to[0,\infty)$ is measurable, bounded, and positive almost everywhere, and $\beta\in(0,2)$ is the covariance exponent. For $u,v\geq0$, the kernel is $q_\beta(u,v)=[u^\beta+v^\beta-(u+v)^\beta]/(1-\beta)$ when $\beta\ne1$. Its continuous extension at $\beta=1$ is $q_1(u,v)=(u+v)\log(u+v)-u\log u-v\log v$, with $0\log0=0$.

Using the Euler approximation, we reconstruct the Gaussian driving increments from observed transitions and use their joint density to obtain a trajectory likelihood. Neural and radial-basis representations model the drift, diffusion, and normalized temporal weight, while a likelihood profile estimates the covariance exponent and diffusion scale. We compare the method with two neural alternatives on the same simulated trajectories in twenty coefficient settings.
\end{abstract}
\maketitle
\begin{quotation}
For candidate drift and diffusion functions, the Euler approximation reconstructs the Gaussian driving increments from the observed transitions. Their joint Gaussian density, computed using the full increment covariance, yields the trajectory likelihood through a change of variables. We develop this estimator for weighted sub-fractional Brownian motion, estimating its temporal weight and covariance exponent alongside the drift and diffusion. Numerical experiments compare three estimation methods using the same observed trajectories.
\end{quotation}

\section{Introduction}\label{sec:introduction}
Stochastic differential equations describe the evolution of a state through a drift term and a noise term whose amplitude is controlled by the diffusion coefficient. Inferring these functions from discrete observations requires a model for the fluctuations that remain after the drift contribution has been removed. If the driving increments are correlated, their covariance also enters this inference problem. Here both the coefficient functions and the noise covariance are unknown, so we estimate them from the same observed trajectories.

Stochastic differential equations driven by Gaussian processes with correlated increments have been used to model animal movement, protein dynamics, and mortality and interest-rate data, among other applications. Because the Gaussian processes involved need not be stationary, we use the separated-increment definition of long-range dependence (LRD) from Refs.~\cite{RamirezMurilloSun2026,Ramirez2026mifOU}. Let $G=(G_t)_{t\geq0}$ be a centered Gaussian process and fix intervals $0<r<v$ and $0<s<t$. The criterion requires an exponent $\kappa\in\mathbb R$ and a finite real constant $C_{r,v;s,t}$ such that
\[
\lim_{u\to\infty}u^{-\kappa}
\operatorname{Cov}(G_v-G_r,G_{u+t}-G_{u+s})
=C_{r,v;s,t},
\]
\[
0<|C_{r,v;s,t}|.
\]

For fractional Brownian motion with Hurst parameter $H\in(0,1)$, $H\ne1/2$, this criterion holds with $\kappa=2H-2$. It also holds for integrated fractional Ornstein--Uhlenbeck processes when $H\ne1/2$. This definition of LRD differs from requiring the absolute increment covariances to be nonsummable: for the integrated fractional Ornstein--Uhlenbeck process, they are nonsummable only when $H>1/2$.

Ram\'irez-Gonz\'alez and Sun\cite{RamirezSun2026ifOU} modeled fin-whale trajectories with an integrated fractional Ornstein--Uhlenbeck process. A later multidimensional extension\cite{Ramirez2026mifOU} introduced correlation between the components of the driving fractional Brownian motion and was applied to bat trajectories. Ram\'irez-Gonz\'alez, Murillo-Salas, and Sun\cite{RamirezMurilloSun2026} analyzed those same bat trajectories using longitude and latitude separately, with independent coordinate processes. In each of the ten coordinate series, a Gaussian model satisfying the LRD criterion above had the lowest Akaike information criterion (AIC). Their exponential-weight model was selected in three series. It is the case $\beta=1$ with $f(t)=c e^{-\lambda t}$, for $c>0$ and $\lambda\geq0$, of the covariance family studied here.

Outside animal telemetry, Kou\cite{Kou2008} used fractional Gaussian noise in a generalized Langevin equation to model fluorescence measurements from single protein molecules. Zhou and Zhou\cite{ZhouZhou2026} used mixed Brownian and fractional Brownian forcing to model mortality and interest-rate data, obtaining Hurst estimates above $1/2$ for both series.

To estimate drift and diffusion functions from discrete observations, a numerical approximation must relate candidate functions to the observed state transitions. Dietrich \textit{et al.}\cite{Dietrich2023} represented both functions by neural networks and trained them jointly. With an Euler--Maruyama approximation, they minimized the Gaussian negative log-likelihood of the one-step transitions; they also derived a loss from the transition density of the Milstein scheme. Fang \textit{et al.}\cite{Fang2022} studied equations driven by symmetric $\alpha$-stable L\'evy noise, $0<\alpha<2$. For $\alpha\ne1$, they first fitted the drift by minimizing the squared Euler prediction error, then held that estimate fixed while fitting the diffusion coefficient by the negative log-likelihood of the one-step $\alpha$-stable transition density. They treated the Cauchy case $\alpha=1$ separately, using absolute prediction error for the drift and a direct estimate of the noise scale.

Ram\'irez-Gonz\'alez and Sun\cite{RamirezChaos2026} used the Tamed--Milstein approximation for equations with Brownian forcing and additive symmetric jumps of finite intensity. They obtained an approximate transition density by inverting its characteristic function and used it to construct a likelihood loss, alongside losses based on conditional moments. Training began with a joint likelihood fit and continued with alternating updates of the drift and diffusion networks. In the present setting, the Gaussian driving increments are correlated across observation times. We therefore evaluate their joint density within each trajectory rather than fitting transitions one at a time.

We apply this approach to equations driven by weighted sub-fractional Brownian motion. Its covariance integrates a kernel indexed by $\beta$ against a temporal weight $f$. The power-weight process arose as a limit of occupation-time fluctuations in branching particle systems,\cite{LopezMimbela2024Occupation} while Ref.~\cite{Ramirez2026WeightedEuler} extended the covariance construction to general admissible weights and studied the resulting process and its Euler approximation. We observe independent state trajectories at common discrete times, with known initial states but unobserved Gaussian drivers. From these trajectories, we estimate the drift $a(x)$, diffusion coefficient $\sigma(x)$, temporal weight $f(t)$, and exponent $\beta$.

For candidate functions $a$ and $\sigma$, the Euler equation reconstructs the driving increments: subtract the proposed drift increment from each observed transition and divide by the proposed diffusion coefficient. A change of variables applied to their joint Gaussian density gives the trajectory likelihood, with covariance determined by $f$ and $\beta$. We normalize $f$ to have unit mean to separate its overall scale from that of $\sigma$.

Our estimator begins with separate fits of the drift and noise components, then refines $a$, $\sigma$, and $f$ jointly, allowing small, constrained curvature corrections to the state-dependent functions. At scheduled stages, it updates $\beta$ by a one-dimensional likelihood search, calculating the optimal global diffusion rescaling for each candidate exponent. The following sections derive the likelihood, specify these fitting steps, and compare three estimation methods using the same simulated trajectories.

\section{The Gaussian process and the Euler observation model}
\label{sec:weighted-theory}

For a fixed $T>0$ and known initial state $x_0\in\mathbb R$, consider
\begin{equation}
 X_t=x_0+\int_0^t a(X_s)\,ds+\int_0^t\sigma(X_s)\,dZ_s^{\beta,f}.
 \label{eq:weighted-continuous}
\end{equation}
The unknown functions $a:\mathbb R\to\mathbb R$ and $\sigma:\mathbb R\to(0,\infty)$ are the drift and diffusion coefficients. The centered Gaussian process $Z^{\beta,f}$ is determined by a temporal weight $f:[0,T]\to[0,\infty)$ and an exponent $\beta\in[0,2]$. We work with measurable, bounded weights that are positive almost everywhere. For a nonnegative integer $k$, $C^k(\mathbb R)$ denotes the functions with $k$ continuous derivatives.

\subsection{Covariance of the Gaussian process}
\label{sec:weighted-covariance}

For $u,v\geq0$ and $0<\beta\leq2$, define
\begin{equation}
 \begin{aligned}
 q_\beta(u,v)&=\frac{u^\beta+v^\beta-(u+v)^\beta}{1-\beta},
          \quad\beta\ne1,\\
 q_1(u,v)&=(u+v)\log(u+v)\\
         &\quad-u\log u-v\log v,
 \end{aligned}
 \label{eq:weighted-q}
\end{equation}
where $0\log0=0$. At $\beta=0$, set $q_0(u,v)=\mathbf1_{\{u>0\}}\mathbf1_{\{v>0\}}$, with $\mathbf1_A$ denoting the indicator of a set $A$. The process starts at zero and has covariance
\begin{equation}
 R_{\beta,f}(s,t)=\int_0^{s\wedge t}f(r)
 q_\beta(s-r,t-r)\,dr .
 \label{eq:weighted-R}
\end{equation}
In Eq.~\eqref{eq:weighted-R}, $f(r)$ weights contributions from the common past, $0\leq r\leq s\wedge t$, while $q_\beta(s-r,t-r)$ determines how they enter the covariance at times $s$ and $t$. The power-weight construction is due to Ref.~\cite{LopezMimbela2024Occupation}; general admissible weights are treated in Ref.~\cite{Ramirez2026WeightedEuler}.

For bounded nonnegative weights, this is a finite covariance for $\beta\in[0,2]$, with the logarithmic continuation at one; its construction and continuity in $\beta$ are established in Ref.~\cite[Theorem~1 and Proposition~8]{Ramirez2026WeightedEuler}. At the endpoints,
\begin{align}
 R_{0,f}(s,t)&=\int_0^{s\wedge t}f(r)\,dr,\label{eq:weighted-beta0}\\
 R_{2,f}(s,t)&=2\int_0^{s\wedge t}f(r)(s-r)(t-r)\,dr.
 \label{eq:weighted-beta2}
\end{align}
The first is the covariance of Brownian motion at the deterministic clock $\int_0^t f(r)\,dr$; the second is that of $\sqrt2\int_0^t(t-r)\sqrt{f(r)}\,dW_r$, where $W$ is standard Brownian motion. For $f>0$ almost everywhere, the covariance of successive increments is strictly positive definite, so their Gaussian density can be evaluated. Appendix~\ref{app:weighted-covariance-proof} gives the short argument for this last property.

\subsection{Euler approximation on the observation grid}

Let $0=t_0<\cdots<t_n=T$ be the observation times and $h_i=t_{i+1}-t_i$. The Euler approximation holds the coefficients at the state preceding each increment:
\begin{equation}
 \begin{aligned}
 X^\pi_{i+1}&=X^\pi_i+h_i a(X^\pi_i)\\
 &\quad+\sigma(X^\pi_i)(Z_{t_{i+1}}-Z_{t_i}),
 \end{aligned}
 \label{eq:weighted-euler}
\end{equation}
with $X^\pi_0=x_0$. Here $\pi=\{t_0,\ldots,t_n\}$ denotes the observation grid and $Z=Z^{\beta,f}$. This recursion is the observation model used to estimate the unknown functions.

Existence of the pathwise solution and convergence of the Euler approximation are established in Ref.~\cite[Theorems~2--4]{Ramirez2026WeightedEuler}. For the autonomous coefficients considered here, those results apply in the regimes $0<\beta<1$ and $1<\beta\leq2$ when $f$ is bounded, $a\in C^1(\mathbb R)$, $\sigma\in C^2(\mathbb R)$, and the coefficients and their indicated derivatives are bounded. The cited construction uses Young integration in the first regime and a random ordinary differential equation in the second. At the Brownian endpoint we use the It\^o convention. The finite-grid likelihood below follows directly from Eq.~\eqref{eq:weighted-euler} under the coefficient assumptions of Proposition~\ref{prop:weighted-density}, including the logarithmic case $\beta=1$ used in the experiments.

\subsection{Covariance of the driving increments on the observation grid}
\label{sec:weighted-grid}

The Euler likelihood requires the joint distribution of the driving increments. Define $e_i=Z_{t_{i+1}}-Z_{t_i}$ for $i=0,\ldots,n-1$ and $e=(e_0,\ldots,e_{n-1})^\top$. Since differences of jointly Gaussian variables remain jointly Gaussian, $e$ has the distribution $\mathcal N_n(0,K(\beta,f))$. The superscript $\top$ denotes transpose. Here $\mathcal N_n(0,K)$ is the centered $n$-dimensional normal distribution with covariance $K$, whose entries are
\begin{equation}
 \begin{aligned}
 K_{ij}(\beta,f)&=R_{\beta,f}(t_{i+1},t_{j+1})\\
 &\quad-R_{\beta,f}(t_{i+1},t_j)\\
 &\quad-R_{\beta,f}(t_i,t_{j+1})+R_{\beta,f}(t_i,t_j).
 \end{aligned}
 \label{eq:weighted-K}
\end{equation}
The diagonal entries are increment variances; the off-diagonal entries describe dependence between increments at different observation times. Their values can depend on the locations of both intervals as well as on their separation.

To calculate this matrix during training, we approximate $f$ by a constant on each of $L$ equal subintervals of $[0,T]$, where $L$ is the number of intervals used for integration. We use $L=24$ in the numerical experiments. Define
\[
I_\ell=\left[\frac{(\ell-1)T}{L},\frac{\ell T}{L}\right),
\qquad \ell=1,\ldots,L,
\]
including the endpoint $T$ in the last interval. On $I_\ell$, replace $f$ by its value at the midpoint:
\[
w_\ell=f\!\left(\frac{(\ell-\tfrac12)T}{L}\right).
\]
The resulting piecewise constant approximation is
\[
f_L(t)=\sum_{\ell=1}^{L}w_\ell\mathbf1_{I_\ell}(t).
\]
Because the covariance depends linearly on $f$, substituting this approximation into its integral gives
\begin{equation}
K(\beta,f_L)=\sum_{\ell=1}^{L}w_\ell B_\ell(\beta),
\qquad
B_\ell(\beta)=K(\beta,\mathbf1_{I_\ell}).
\label{eq:weighted-cell-K}
\end{equation}

Each $B_\ell(\beta)$ is an $n\times n$ increment covariance matrix obtained with weight one on $I_\ell$ and zero elsewhere. Appendix~\ref{app:weighted-cell} gives its analytic entries, including at $\beta=1$. For a fixed exponent, only the network values $w_\ell$ need updating in this matrix sum. We can therefore update $K$ through this weighted sum instead of integrating the changing network output at every step, while retaining every observed increment and all pairwise covariances.

The number $L$ controls the approximation of the covariance integral, whereas $n$ is fixed by the observations. For fixed $\beta$, observation times, and a continuous weight $f$, the midpoint approximation converges uniformly. Indeed, every point in a cell lies within $T/(2L)$ of its midpoint, and continuity on the compact interval $[0,T]$ gives
\[
 \begin{aligned}
 \sup_{t\in[0,T]}|f_L(t)-f(t)|
 &\leq \sup_{\substack{u,v\in[0,T]\\|u-v|\leq T/(2L)}}
 |f(u)-f(v)|\\
 &\longrightarrow0.
 \end{aligned}
\]
Substitution into the covariance integral then gives entrywise convergence of $K(\beta,f_L)$ to $K(\beta,f)$. When the limiting matrix is positive definite, its inverse, Cholesky factor, and log determinant also converge, as does the likelihood at fixed coefficient functions. This justifies refinement of the covariance calculation for a fixed continuous candidate; it does not assert convergence of estimates obtained by retraining as $L$ increases.

\subsection{Likelihood of an observed Euler trajectory}
\label{sec:weighted-likelihood}

Suppose that $P$ trajectories are observed without measurement error at common times $t_0,\ldots,t_n$, with known initial states. Let $x_i^{(p)}$ denote the state of trajectory $p$ at time $t_i$, and write $h_i=t_{i+1}-t_i$. All paths share the same coefficients and covariance parameters, collected in $\theta=(a,\sigma,\beta,f)$, and are driven by independent copies of the Gaussian process.

For a candidate $\theta$, the Euler equation gives the driving increments as functions of the observed states:
\begin{equation}
 r_i^{(p)}(\theta)=
 \frac{x_{i+1}^{(p)}-x_i^{(p)}-h_i a(x_i^{(p)})}
      {\sigma(x_i^{(p)})},
 \qquad i=0,\ldots,n-1.
 \label{eq:weighted-residual}
\end{equation}
For a trajectory generated by the candidate Euler model, the reconstructed increments form a centered Gaussian vector with covariance $K=K(\beta,f)$. We evaluate this model density at the increments reconstructed from the observed states. Assuming that $K$ is strictly positive definite, written $K\succ0$, its density is
\[
\varphi_K(z)=
\frac{\exp\!\left(-\tfrac12 z^\top K^{-1}z\right)}
     {(2\pi)^{n/2}(\det K)^{1/2}},
\qquad z\in\mathbb R^n.
\]

For a fixed initial state, the Euler recursion gives an invertible transformation between the state vector $(x_1^{(p)},\ldots,x_n^{(p)})$ and the increment vector
$r^{(p)}(\theta)=(r_0^{(p)}(\theta),\ldots,r_{n-1}^{(p)}(\theta))^\top$.
If $a$ and $\sigma$ are continuously differentiable and $\sigma>0$, this transformation and its inverse are continuously differentiable. Since each $r_i^{(p)}$ depends only on $x_i^{(p)}$ and $x_{i+1}^{(p)}$, the state-to-increment Jacobian is lower triangular, with diagonal entries
\[
\frac{\partial r_i^{(p)}}{\partial x_{i+1}^{(p)}}
=\frac{1}{\sigma(x_i^{(p)})}.
\]
Its determinant is therefore
$\prod_{i=0}^{n-1}\sigma(x_i^{(p)})^{-1}$.

Let $p_\theta(\,\cdot\mid x_0^{(p)})$ denote the conditional joint density of the subsequent states under this Euler model. The change-of-variables formula gives
\[
p_\theta\!\left(x_1^{(p)},\ldots,x_n^{(p)}
\mid x_0^{(p)}\right)
=
\varphi_K\!\left(r^{(p)}(\theta)\right)
\prod_{i=0}^{n-1}\frac{1}{\sigma(x_i^{(p)})}.
\]
The likelihood for the $P$ trajectories is the product of these conditional densities, by independence of the driving processes.
\begin{proposition}[Negative log likelihood of the Euler model]
\label{prop:weighted-density}
Assume that $a$ and $\sigma$ are continuously differentiable, $\sigma(x)>0$ for every $x$, and $K=K(\beta,f)\succ0$. For $P$ independent trajectories with known initial states, the negative log likelihood divided by $nP$, after omitting the constant $\frac12\log(2\pi)$, is
\begin{equation}
 \begin{aligned}
 \mathcal L(\theta)&=\frac{\log\det K}{2n}\\
 &\quad+\frac1{nP}\sum_{p=1}^P\sum_{i=0}^{n-1}
       \log\sigma(x_i^{(p)})\\
 &\quad+\frac1{2nP}\sum_{p=1}^P
       r^{(p)\top}K^{-1}r^{(p)},
 \end{aligned}
 \label{eq:weighted-likelihood}
\end{equation}
where $r^{(p)}=r^{(p)}(\theta)$ is defined by
Eq.~\eqref{eq:weighted-residual}.
\end{proposition}

The three terms follow directly from the trajectory density: the log determinant comes from the Gaussian normalizing factor, the sum of log diffusion coefficients comes from the Jacobian, and the quadratic form measures the reconstructed increments using their joint covariance. Both $a$ and $\sigma$ enter the reconstructed increments; $\sigma$ also enters the Jacobian determinant. The temporal weight $f$ and exponent $\beta$ determine $K$. Dividing by the total number of transitions, $nP$, changes the scale of the loss but not its minimizer.

For numerical evaluation, use the Cholesky factorization
$K=L_KL_K^\top$, where $L_K$ is lower triangular with positive diagonal, and solve
\[
L_Kz^{(p)}=r^{(p)}.
\]
Then
\[
r^{(p)\top}K^{-1}r^{(p)}=\|z^{(p)}\|_2^2,
\qquad
\|z\|_2^2=\sum_{j=1}^{n}z_j^2.
\]
Thus, the quadratic term can be evaluated by triangular solves without explicitly forming $K^{-1}$.

The derivation applies to any centered Gaussian driver with positive definite increment covariance. The weighted family specifies this covariance through $f$ and $\beta$. Equation~\eqref{eq:weighted-likelihood} is the exact normalized negative log likelihood for the specified Euler model, up to the omitted constant. When Euler discretization approximates the continuous-time equation, or when $K(\beta,f_L)$ approximates the continuous-weight covariance, we use the resulting criterion as a Gaussian quasi-likelihood. The Gaussian law is imposed on the driving increments; the state vector need not be Gaussian.

Temporal dependence also affects the conditional distribution of each transition. For one path, write $X_i=X_i^\pi$ and $\Delta X_i=X_{i+1}-X_i$. Given $\theta$ and $X_0$, the Euler transformation makes the history $(X_0,\ldots,X_i)$ equivalent to the earlier reconstructed increments
\[
r_{<i}=(r_0,\ldots,r_{i-1})^\top.
\]
For $i\geq1$, define their covariance matrix and their covariance vector with the next increment by
\[
K_{<i,<i}=(K_{jk})_{j,k=0}^{i-1},
\qquad
k_i=(K_{0i},\ldots,K_{i-1,i})^\top.
\]
Let $\mathbb E_\theta$ and $\operatorname{Var}_\theta$ denote expectation and variance under the candidate Euler model. Applying the Gaussian conditioning formula to the driving increments gives
\begin{align}
 \mathbb E_\theta[\Delta X_i\mid X_0,\ldots,X_i]
 &=h_i a(X_i)
   +\sigma(X_i)k_i^\top K_{<i,<i}^{-1}r_{<i},
 \label{eq:weighted-conditional-mean}\\
 \operatorname{Var}_\theta[\Delta X_i\mid X_0,\ldots,X_i]
 &=\sigma(X_i)^2
   \bigl(K_{ii}-k_i^\top K_{<i,<i}^{-1}k_i\bigr).
 \label{eq:weighted-conditional-var}
\end{align}
For $i=0$, these expressions reduce to $h_0a(X_0)$ and
$\sigma(X_0)^2K_{00}$, respectively.

The conditional mean therefore combines the drift contribution with a term from the past driving increments. A regression of the observed increments on the current state alone need not isolate the drift. For this reason, the squared-residual fit used below serves as an initialization and is followed by likelihood fitting with the full covariance. Replacing $K$ by its diagonal would instead impose independence between the driving increments.

\subsection{Separating the diffusion scale from the temporal weight}
\label{sec:weighted-identification}

The overall scales of $f$ and $\sigma$ cannot be estimated separately, because a change in the covariance scale can be offset by a change in diffusion.

\begin{proposition}[Scale equivalence]
\label{prop:weighted-scale}
For any $c>0$, replacing $(f,\sigma)$ by $(cf,\sigma/\sqrt c)$, with $a$ and $\beta$ unchanged, preserves the law of every Euler trajectory and the loss in Eq.~\eqref{eq:weighted-likelihood}. We fix this constant scale ambiguity by imposing
\begin{equation}
 \frac1T\int_0^T f(t)\,dt=1,\qquad \sigma(x)>0.
 \label{eq:weighted-scale-convention}
\end{equation}
The unit-mean condition fixes the scale, while positivity excludes the equivalent global sign change of $\sigma$.
\end{proposition}

Indeed, linearity of the covariance in $f$ gives
$K(\beta,cf)=cK(\beta,f)$, while division of $\sigma$ by
$\sqrt c$ multiplies every reconstructed increment by $\sqrt c$.
The quadratic term in the loss is therefore unchanged. The
log-determinant term increases by $\frac12\log c$, and the
log-diffusion term decreases by the same amount.

In the numerical implementation, normalization uses the same cells
as the covariance calculation. Let $v:[0,T]\to(0,\infty)$ be the
continuous positive network output before normalization. For cell
$I_\ell$, write $|I_\ell|$ for its length and
$m_\ell=(\ell-\tfrac12)T/L$ for its midpoint. Define
\[
 D_L=\frac1T\sum_{\ell=1}^{L}|I_\ell|v(m_\ell),
 \qquad w_\ell=\frac{v(m_\ell)}{D_L}.
\]
For equal cells, the divisor $D_L>0$ is the arithmetic mean of the
midpoint values and is recomputed whenever the network changes.
The reported continuous curve is $v(t)/D_L$, and each $w_\ell$
is exactly this curve evaluated at $m_\ell$.

To distinguish this curve from the step function entering the
covariance, set
\[
 v_L(t)=\sum_{\ell=1}^{L}v(m_\ell)\mathbf1_{I_\ell}(t),
 \qquad f_L(t)=\frac{v_L(t)}{D_L}.
\]
Thus $f_L$ has cell values $w_\ell$, and its normalization is exact:
\[
 \frac1T\int_0^T f_L(t)\,dt
 =\frac{T^{-1}\sum_{\ell=1}^{L}|I_\ell|v(m_\ell)}{D_L}=1.
\]

The dependence of $D_L$ on $L$ must also be accounted for when
refining the integration cells. Hold the network output $v$ fixed
and define its integral mean and normalized continuous weight by
\[
 D=\frac1T\int_0^T v(t)\,dt>0,
 \qquad f(t)=\frac{v(t)}{D}.
\]
With $\|g\|_\infty=\sup_{t\in[0,T]}|g(t)|$, the midpoint argument
in Sec.~\ref{sec:weighted-grid} gives
$\varepsilon_L=\|v_L-v\|_\infty\to0$. Since
\[
 |D_L-D|
 =\left|\frac1T\int_0^T[v_L(t)-v(t)]\,dt\right|
 \leq\varepsilon_L,
\]
we also have $D_L\to D>0$. Consequently,
\[
 \begin{aligned}
 \|f_L-f\|_\infty
 &\leq\frac{\varepsilon_L}{D_L}
 +\frac{\|v\|_\infty|D_L-D|}{D_LD}\\
 &\longrightarrow0.
 \end{aligned}
\]
The covariance convergence in Sec.~\ref{sec:weighted-grid} therefore
holds for the normalized weights as well. The reported continuous
curve $v/D_L$ converges uniformly to $f=v/D$; at finite $L$, its
integral mean is $D/D_L$, which need not equal one. This argument
keeps the network parameters, exponent, and observation times fixed;
it establishes convergence of the covariance approximation, not of
estimates obtained by retraining as $L$ changes.

The exponential-weight model cited in the introduction gives an
explicit example of the scale normalization. In
Ref.~\cite[Eq.~(11)]{RamirezMurilloSun2026}, the weight
$A e^{-\lambda t}$ appears with a factor two in the covariance
integral, where $A>0$ is the covariance scale and $\lambda\geq0$
is the decay rate. Under our convention in
Eq.~\eqref{eq:weighted-R}, the same model has $\beta=1$, zero
drift, $\sigma=1$, and weight $2A e^{-\lambda t}$. Define the
mean of the exponential over the observation interval by
\begin{equation}
 \begin{aligned}
 m_T(\lambda)
 &=\frac1T\int_0^T e^{-\lambda u}\,du\\
 &=\begin{cases}
 \dfrac{1-e^{-\lambda T}}{\lambda T},&\lambda>0,\\[4pt]
 1,&\lambda=0.
 \end{cases}
 \end{aligned}
 \label{eq:telemetry-exponential-mean}
\end{equation}
The original weight has mean $2A m_T(\lambda)$. Dividing by this
mean and multiplying the original diffusion coefficient by its
square root gives
\begin{equation}
 \begin{aligned}
 \beta&=1,\qquad a(x)=0,\\
 f(t)&=\frac{e^{-\lambda t}}{m_T(\lambda)},\qquad
 \sigma(x)=\sqrt{2A m_T(\lambda)}.
 \end{aligned}
 \label{eq:telemetry-special-case}
\end{equation}
The normalized weight satisfies the required condition directly:
\[
 \frac1T\int_0^T f(t)\,dt
 =\frac{T^{-1}\int_0^T e^{-\lambda t}\,dt}{m_T(\lambda)}=1.
\]
Its integral is $T$; its integral mean is one. The change in
diffusion preserves the original covariance because
\[
 \sigma^2 f(t)
 =2A m_T(\lambda)\frac{e^{-\lambda t}}{m_T(\lambda)}
 =2A e^{-\lambda t}.
\]
Indeed, with zero drift and constant diffusion, the state
covariance is $\sigma^2 R_{1,f}$, and $R_{1,f}$ is linear in $f$.
For the numerical covariance, replace $m_T(\lambda)$ in both
$f$ and $\sigma$ by
$m_{T,L}(\lambda)=T^{-1}\sum_{\ell=1}^{L}|I_\ell|e^{-\lambda m_\ell}$.
The resulting cell weights have unit mean and satisfy
$\sigma^2w_\ell=2A e^{-\lambda m_\ell}$, preserving the
corresponding midpoint covariance exactly.

Fixing the scale does not establish uniqueness of the temporal
weight. On a fixed observation grid, the likelihood depends on
$f$ only through the finitely many entries of $K(\beta,f)$.
Different positive weights with unit mean can produce the same
matrix; Appendix~\ref{app:weighted-identification-proof} gives an
example.

The methods in Sec.~\ref{sec:methods} estimate the functions within specified representations that enforce positivity and normalization. Independent paths provide repeated observations on the common time grid, and the use of several initial states gives information across different state regions. Neither these design choices nor the normalization alone establish uniqueness within the chosen representations.

\section{Estimation using the Gaussian trajectory likelihood}
\label{sec:methods}

We consider three methods for estimating $a$, $\sigma$, $f$, and $\beta$ from the Gaussian trajectory likelihood in Eq.~\eqref{eq:weighted-likelihood}. They differ in their function representations and optimization schedules.

Method 1 (M1) updates the parameters representing $a$, $\sigma$, and $f$ by gradient steps. At selected training cycles, it also updates $\beta$ and the global diffusion scale while holding the function parameters fixed. For each candidate exponent, the multiplier of the current diffusion curve that minimizes the Gaussian loss is obtained analytically. Substituting this multiplier into the loss leaves a one-dimensional optimization over $\beta$. Training begins by initializing the drift, continues by fitting diffusion and the temporal weight with the drift fixed, and then refines all three functions jointly. Figure~\ref{fig:architecture} summarizes the function representations, fitting objective, and sequence of updates used by M1.

Methods 2 and 3 (M2 and M3) represent each of the three functions by a neural network and use Adam to update the network parameters and a transformed scalar parameter for $\beta$. M2 updates all four components from the first cycle. M3 first initializes the drift, then fits diffusion, the temporal weight, and the exponent with the drift fixed, before proceeding to joint updates.

For the numerical formulas below, write $X_{i,p}=x_i^{(p)}$ for the observed state and $Y_{i,p}=X_{i+1,p}-X_{i,p}$ for its increment. The reconstructed driving increments from Eq.~\eqref{eq:weighted-residual} are denoted by $r_{i,p}=r_i^{(p)}$, with vector $\boldsymbol r_p=(r_{0,p},\ldots,r_{n-1,p})^\top$ on path $p$. Here $p=1,\ldots,P$, $i=0,\ldots,n-1$, and $N=Pn$ is the total number of observed transitions.

\subsection{Data used in each loss evaluation}
\label{sec:m-data-evaluation}

In all three methods, every Adam step evaluates its loss and
gradient using all $P$ trajectories and all $n$ transitions within
each trajectory. No minibatches are selected, and paths are not
divided into temporal windows.
To evaluate the Gaussian loss, we evaluate the drift and diffusion
representations at each of the $N$ left-endpoint states $X_{i,p}$,
using the input transformations specified below. The next states
enter through the increments $Y_{i,p}$. These evaluations produce
one complete reconstructed increment vector $\boldsymbol r_p$ of
length $n$ for each path. The coefficient representations receive
individual state inputs; the likelihood combines their outputs
across the full trajectories.

The temporal network receives the integration-cell midpoints.
Its normalized outputs and the current exponent determine a common
$n\times n$ covariance matrix $K$, used for every path. We evaluate
each complete vector $\boldsymbol r_p$ with this full covariance,
sum the conditional trajectory negative log densities, including
the change-of-variables terms, and divide by $N$, as in
Eq.~\eqref{eq:weighted-likelihood}. Thus the
loss retains temporal correlations within each path and combines
paths through their independence. Appendix~\ref{app:full-data-evaluation}
describes the Cholesky calculation used in the code.

M1 adds its diffusion penalty once to this normalized likelihood;
M2 and M3 use the unpenalized likelihood. The drift initializations
of M1 and M3 also use all $N$ transitions, through the
squared-residual loss defined below. The integration cells and the
points used to measure curvature are auxiliary numerical grids,
not training batches.

\begin{figure*}[tbp]
\centering\manuscriptfigure[width=.92\textwidth]{architecture.pdf}
\caption{Method 1. (A) The observed state enters the drift and diffusion representations, and time enters the temporal-weight network. The Gaussian basis has fixed centers and width; its coefficients are learned. Each Adam step uses all observed trajectories without temporal subdivision. (B) The drift and diffusion coefficient reconstruct the driving increments; the temporal weight and exponent determine the increment covariance used in the objective $\mathcal J$ of Eq.~\eqref{eq:m-objective}. (C) Cycles 1--5 initialize the drift using squared increment residuals. Cycles 6--10 fit the noise with the drift fixed. Cycles 11--510 refine the three functions jointly, test small curvature corrections, and take three additional diffusion updates. The exponent and global diffusion scale are updated after cycles 5, 10, and every tenth cycle thereafter. (D) Each curvature correction is constrained by the preceding joint update, as detailed in Sec.~\ref{sec:m-curvature}.}
\label{fig:architecture}
\end{figure*}

\subsection{Representing the functions in Method 1}
\label{sec:m-networks}

The representations must allow a signed drift, a positive diffusion coefficient, and a positive temporal weight with the normalization specified in Sec.~\ref{sec:weighted-identification}. We use a neural network for the drift and another for the temporal weight. Diffusion is obtained by applying a positive transformation to a fixed smooth basis expansion with fitted coefficients. This representation permits a quadratic penalty on the second derivative of the expansion before that transformation. 

\subsubsection{State input and drift}

The state functions are evaluated at the start of each transition. We therefore compute the mean and sample standard deviation over all $N$ left-endpoint states:
\[
\begin{aligned}
\bar x
&=\frac1N\sum_{p=1}^{P}\sum_{i=0}^{n-1}X_{i,p},\\
\operatorname{sd}(X_{i,p})
&=\sqrt{\frac1{N-1}
\sum_{p=1}^{P}\sum_{i=0}^{n-1}(X_{i,p}-\bar x)^2}.
\end{aligned}
\]
We define the fixed scale $s_x$ and standardized input $z(x)$ by
\begin{equation}
s_x=\max\{\operatorname{sd}(X_{i,p}),10^{-6}\},
\qquad
z(x)=\frac{x-\bar x}{s_x}.
\label{eq:m-standardization}
\end{equation}
The lower bound prevents division by a very small standard deviation. Both $\bar x$ and $s_x$ remain fixed during and after fitting.

Let $\boldsymbol\theta_a$ collect the weights and biases of a fully connected network $A_{\boldsymbol\theta_a}$ with four hidden layers of 32 exponential linear units (ELUs). Each ELU applies $u$ for $u>0$ and $e^u-1$ otherwise.\cite{Clevert2016} The output layer is affine, with no activation, so the drift can have either sign. The network receives the transformed input $\tanh z(x)$ and returns the drift in its original units:
\begin{equation}
a_{\boldsymbol\theta_a}(x)
=A_{\boldsymbol\theta_a}(\tanh z(x)).
\label{eq:m-drift-net}
\end{equation}

\subsubsection{Diffusion coefficient}
We first construct a Gaussian radial basis expansion with an affine term,\cite{Fasshauer2007} and then transform its output to obtain a positive diffusion coefficient. The affine term provides a level and a linear trend, while the Gaussian basis functions allow nonlinear departures from that trend.

Let $z_-$ and $z_+$ denote the smallest and largest standardized states at the start of the observed transitions:
\[
z_-=\min_{i,p}z(X_{i,p}),
\qquad
z_+=\max_{i,p}z(X_{i,p}),
\]
where $i=0,\ldots,n-1$ and $p=1,\ldots,P$. For the nondegenerate range $z_-<z_+$, we place sixteen Gaussian basis functions at equally spaced centers $c_j$, including both endpoints, and assign them a common width $b_\sigma$:
\[
\begin{aligned}
c_j&=z_-+\frac{j-1}{15}(z_+-z_-),
\qquad j=1,\ldots,16,\\
b_\sigma&=1.25\frac{z_+-z_-}{15}.
\end{aligned}
\]
Thus the width is 1.25 times the distance between consecutive centers. The centers and width are calculated from the observations and remain fixed during training. The number of basis functions and the width factor are the same in all experiments.

Let $\boldsymbol q=(q_0,\ldots,q_{17})^{\mathsf T}\in\mathbb R^{18}$ collect the coefficients to be fitted. The real-valued expansion is
\begin{equation}
g_{\boldsymbol q}(z)
=
q_0+q_1z+
\sum_{j=1}^{16}q_{j+1}
\exp\left[-\frac{(z-c_j)^2}{2b_\sigma^2}\right].
\label{eq:m-rbf}
\end{equation}
Here $q_0$ and $q_1$ determine the affine component, and the remaining sixteen coefficients weight the Gaussian basis functions. These coefficients are unrestricted in sign.

To obtain a positive diffusion coefficient, define
$\operatorname{softplus}(u)=\log(1+e^u)$ and let $s>0$ be a global scale parameter. We set
\begin{equation}
\sigma_{\boldsymbol q,s}(x)
=
s\left\{
\operatorname{softplus}\bigl(g_{\boldsymbol q}(z(x))\bigr)
+10^{-8}
\right\}.
\label{eq:m-sigma-net}
\end{equation}
The offset adds a small positive floor before multiplication by $s$. The coefficients $\boldsymbol q$ control the shape and level of the curve, whereas $s$ rescales the entire diffusion function with $\boldsymbol q$ fixed. This separate scale permits the analytic diffusion rescaling used in the exponent profile described in Sec.~\ref{sec:m-profile}. We refer to $g_{\boldsymbol q}$, the function before softplus, as the diffusion preactivation.

\subsubsection{Temporal weight and increment covariance}

We represent the temporal weight by a neural network and normalize its positive output before using it to construct the increment covariance. Let $\boldsymbol\eta$ collect the weights and biases of a fully connected network $F_{\boldsymbol\eta}$ with three hidden layers of 48 ELU units and an affine output. The network receives the transformed time $2t/T-1$, which maps $[0,T]$ to $[-1,1]$. Applying softplus to its output gives a positive function, denoted by $v_{\boldsymbol\eta}$.

Normalization uses the $L=24$ equal integration cells $I_\ell$ from Sec.~\ref{sec:weighted-grid}. Their lengths are $\Delta_\ell=T/L$, and their midpoints are $m_\ell=(\ell-\tfrac12)T/L$. For this fixed partition, write $D_{\boldsymbol\eta}$ for the midpoint mean of $v_{\boldsymbol\eta}$ and $f_{\boldsymbol\eta}$ for the normalized curve:
\begin{align}
v_{\boldsymbol\eta}(t)
&=\operatorname{softplus}
  \bigl(F_{\boldsymbol\eta}(2t/T-1)\bigr),
\nonumber\\
D_{\boldsymbol\eta}
&=\sum_{\ell=1}^{L}\frac{\Delta_\ell}{T}
  v_{\boldsymbol\eta}(m_\ell)
 =\frac1L\sum_{\ell=1}^{L}v_{\boldsymbol\eta}(m_\ell),
\nonumber\\
f_{\boldsymbol\eta}(t)
&=\frac{v_{\boldsymbol\eta}(t)}{D_{\boldsymbol\eta}}.
\label{eq:m-weight-net}
\end{align}
Thus $D_{\boldsymbol\eta}$ is the divisor $D_L$ from Sec.~\ref{sec:weighted-identification}, evaluated at the current network output. It is recomputed whenever $\boldsymbol\eta$ changes, ensuring
\[
\sum_{\ell=1}^{L}\frac{\Delta_\ell}{T}
f_{\boldsymbol\eta}(m_\ell)=1.
\]
This normalization fixes the scale of the temporal weight relative to diffusion.

For each cell $I_\ell$, let $B_\ell(\beta)$ be its $n\times n$ increment covariance contribution, defined in Eq.~\eqref{eq:weighted-cell-K}. The numerical covariance assigns the normalized midpoint value $f_{\boldsymbol\eta}(m_\ell)$ to that cell:
\begin{equation}
K_{\beta,\boldsymbol\eta}
=
\sum_{\ell=1}^{L}
f_{\boldsymbol\eta}(m_\ell)B_\ell(\beta).
\label{eq:m-product-K}
\end{equation}
This is the covariance obtained from the piecewise constant weight whose value on $I_\ell$ is $f_{\boldsymbol\eta}(m_\ell)$. For fixed $\beta$, updating the temporal network changes the coefficients of this matrix sum; changing $\beta$ also changes the cell matrices.

The parameters $\boldsymbol\eta$ are fitted through the Gaussian trajectory loss in Eq.~\eqref{eq:weighted-likelihood}. Its gradient passes through the covariance sum and the normalized network outputs, including the dependence of $D_{\boldsymbol\eta}$ on $\boldsymbol\eta$. Each update therefore accounts for the effect of the shared denominator on all cell weights.

At the end of training, let $\widehat{\boldsymbol\eta}$ denote the fitted network weights and biases. The reported continuous estimate is
\[
\widehat f(t)
=
\frac{v_{\widehat{\boldsymbol\eta}}(t)}
     {D_{\widehat{\boldsymbol\eta}}},
\]
and the fitted covariance uses $w_\ell=\widehat f(m_\ell)$. As established in Sec.~\ref{sec:weighted-identification}, the step function entering the covariance has integral mean exactly one, whereas the integral mean of the reported continuous curve may differ from one by the midpoint approximation error.

\subsection{The objective for fitting the functions}
\label{sec:m-objective}

After the drift initialization described below, M1 fits the function
parameters using the Gaussian negative log likelihood
$\mathcal L$ in Eq.~\eqref{eq:weighted-likelihood}, together with a
penalty on the diffusion representation. The current drift and
diffusion determine the reconstructed increments, while the temporal
weight and exponent determine their covariance through
Eq.~\eqref{eq:m-product-K}. The state representations are continuously
differentiable and satisfy $\sigma>0$, as required by the coefficient
assumptions in Proposition~\ref{prop:weighted-density}.

Flexible representations can introduce oscillations into the estimated
coefficients when fitting a finite noisy sample. Roughness penalties
control this variability by balancing fit to the observations against
variation in the estimated function. Eilers and Marx\cite{EilersMarx1996}
discuss the classical integrated squared second-derivative penalty and
its relation to discrete difference penalties in spline smoothing.
We use this principle to regularize the diffusion representation.

The penalty discourages rapid changes of slope in the diffusion
preactivation $g_{\boldsymbol q}$. Its second derivative, taken with
respect to the standardized state $z$, measures how the slope changes.
Squaring this derivative penalizes changes in either direction without
allowing positive and negative values to cancel. An affine function
has zero second derivative, so this measure does not penalize a
constant level or a linear trend. Here curvature is measured through
these changes of slope, before applying the positive transformation
that defines $\sigma$.

The implementation evaluates $g_{\boldsymbol q}''$ analytically at
$J=129$ equally spaced points. Define the nodes $z_k$ and normalized
trapezoidal weights $\omega_k$ by
\begin{equation}
 \begin{aligned}
 z_k&=z_-+\frac{k-1}{J-1}(z_+-z_-),\quad k=1,\ldots,J,\\
 \omega_1&=\omega_J=\frac{1}{2(J-1)},\\
 \omega_k&=\frac{1}{J-1},\quad k=2,\ldots,J-1.
 \end{aligned}
 \label{eq:m-penalty-grid}
\end{equation}
Let $\Omega_\sigma$ be the fixed positive semidefinite $18\times18$
matrix constructed in Eq.~\eqref{eq:app-roughness}. It satisfies the
exact identity
\begin{equation}
 \boldsymbol q^{\mathsf T}\Omega_\sigma\boldsymbol q
 =\sum_{k=1}^{J}\omega_k[g_{\boldsymbol q}''(z_k)]^2
 +10^{-8}\sum_{j=1}^{16}q_{j+1}^2.
 \label{eq:m-discrete-roughness}
\end{equation}
The first term is the composite trapezoidal rule for the integral
mean of $[g_{\boldsymbol q}'']^2$ over $[z_-,z_+]$. The second adds a
small quadratic penalty on the sixteen Gaussian coefficients.
Neither term involves $q_0$ or $q_1$. The matrix depends only on the
fixed basis and evaluation grid, so it is computed once. Appendix~\ref{app:diffusion-penalty}
gives its construction, quantifies the quadrature error, and shows
how the corresponding integral controls changes in slope.

The function updates use the penalized objective
\begin{equation}
 \mathcal J
 =\mathcal L
 +\frac{10^{-3}}{2}
  \boldsymbol q^{\mathsf T}\Omega_\sigma\boldsymbol q,
 \label{eq:m-objective}
\end{equation}
where $10^{-3}$ is the fixed penalty coefficient. The objective uses
the finite sum in Eq.~\eqref{eq:m-discrete-roughness}; it does not
evaluate the curvature integral during training. This penalty acts
on $g_{\boldsymbol q}$, rather than directly on the second derivative
of $\sigma$. No fixed drift penalty is added to $\mathcal J$. The
separate local corrections to the drift and diffusion curves are
described in Sec.~\ref{sec:m-curvature}.

During gradient updates based on $\mathcal J$, we hold $\beta$ and
the global diffusion multiplier $s$ fixed and update the function
parameters selected for the current training stage. Differentiation
includes the dependence of the reconstructed increments on $a$ and
the full positive function $\sigma$, as well as the log-Jacobian
term involving $\sigma$. For the temporal network, it includes both
the covariance weights and their shared normalizing denominator
$D_{\boldsymbol\eta}$. Appendix~\ref{app:implementation} gives these
derivatives and their propagation to the representation parameters.

\subsection{Method 1: the sequence of estimation steps}
\label{sec:m1-training}

M1 has three stages. We first initialize the drift by minimizing
squared Euler increment residuals. We then fit the diffusion
coefficient and temporal weight using the Gaussian likelihood,
with the drift held fixed. Finally, we update all three functions
jointly. The initial squared-residual criterion does not account
for temporal correlations in the driving increments. It can
therefore attribute part of the noise contribution in the
conditional mean of Eq.~\eqref{eq:weighted-conditional-mean}
to the drift. Allowing the drift to change during the joint
likelihood fit lets us revise this initial estimate while
accounting for the full increment covariance.

Before these stages, we initialize the drift and diffusion as
constant functions whose levels are calculated from the
observations, and set $f=1$, $\beta=1$, and $s=1$.
Appendix~\ref{app:implementation} specifies these initial values.
Function parameters are updated using Adam,\cite{KingmaBa2015}
with a learning rate of $10^{-3}$ and the complete-data evaluation
in Sec.~\ref{sec:m-data-evaluation}. A training cycle is one
execution of the operations assigned to a stage. It may contain
several Adam steps, each using all observed transitions again;
it is not a sequence of updates over disjoint data batches.
The subsections below specify these operations and the scheduled
updates of $\beta$ and $s$, in the order shown in
Fig.~\ref{fig:architecture}(C).

\subsubsection{Cycles 1--5: obtain a starting drift}

During the first five cycles, we update only the drift parameters
$\boldsymbol\theta_a$. We compare the observed increment $Y_{i,p}$
with the deterministic Euler contribution $h_i a(X_{i,p})$.
To normalize these differences, we compute the fixed scale
$s_Y=\max\{\operatorname{sd}(Y_{i,p}),10^{-8}\}$ from all
$N=Pn$ observed increments, using the sample standard deviation
with denominator $N-1$. Each cycle takes one Adam step using
the loss
\begin{equation}
 \mathcal L_a
 =\frac1N\sum_{p=1}^{P}\sum_{i=0}^{n-1}
 \left[
 \frac{Y_{i,p}-h_i a(X_{i,p})}{s_Y}
 \right]^2.
 \label{eq:m-warmup}
\end{equation}
The scale $s_Y$ remains fixed and is distinct from the diffusion
multiplier $s$. These five updates provide a starting drift for
the subsequent likelihood fit. Throughout these updates,
$\boldsymbol q$ and $\boldsymbol\eta$ retain their initial values,
and $\beta=s=1$.

After cycle 5, we perform the first joint update of $\beta$ and $s$,
holding $\boldsymbol\theta_a$, $\boldsymbol q$, and
$\boldsymbol\eta$ fixed. For each candidate exponent, we calculate
analytically the diffusion multiplier that minimizes the Gaussian
loss $\mathcal L$. We then compare candidate exponents using the
loss evaluated at their corresponding optimal multipliers.
Thus the numerical search is only over $\beta$; the associated
value of $s$ is obtained analytically. Section~\ref{sec:m-profile}
derives this update and specifies the search procedure.

\subsubsection{Cycles 6--10: fit the noise with the drift fixed}

With the initial drift held fixed, we fit the diffusion coefficient
and temporal weight using the penalized Gaussian objective
$\mathcal J$. Each cycle starts with one joint Adam step on
$\boldsymbol q$ and $\boldsymbol\eta$, updating both the diffusion
representation and the temporal weight that determines the
increment covariance. We then hold $\boldsymbol\eta$ fixed and
take three additional Adam steps on $\boldsymbol q$, using the
same objective. These additional steps allow the diffusion
coefficient to adjust to the updated covariance. Each cycle
therefore contains four updates of $\boldsymbol q$ and one of
$\boldsymbol\eta$; the drift parameters remain unchanged. Each
of these four Adam steps reevaluates $\mathcal J$ and its gradient
on the complete data set at the current parameters.

Throughout cycles 6--10, $\beta$ and $s$ retain the values obtained
after cycle 5. After the final diffusion update in cycle 10,
we hold $\boldsymbol\theta_a$, $\boldsymbol q$, and
$\boldsymbol\eta$ fixed and repeat the exponent--scale profile
over candidate $\beta$ values, as described in Sec.~\ref{sec:m-profile}.
\subsubsection{Cycles 11--510: refine all three functions}

Starting with cycle 11, each cycle begins with one joint Adam
step on $\boldsymbol\theta_a$, $\boldsymbol q$, and
$\boldsymbol\eta$, using the penalized objective $\mathcal J$
with $\beta$ and $s$ fixed. The drift is therefore updated
through the Gaussian likelihood, allowing its initial estimate
to change in response to the fitted noise covariance.

If this joint step decreases $\mathcal J$, we test a local
curvature correction to the drift, followed by one to diffusion.
The drift correction changes only $\boldsymbol\theta_a$;
the diffusion correction changes only $\boldsymbol q$.
Each correction must reduce the discrete curvature measure
defined in Sec.~\ref{sec:m-curvature}, satisfy bounds on its
parameter and curve displacements, and keep any increase in
$\mathcal J$ within a specified fraction of the decrease
obtained by Adam. If no candidate satisfies these conditions,
the corresponding function retains its post-Adam parameters.
If the joint step does not decrease $\mathcal J$, both
corrections are skipped.

We then take three additional Adam steps on $\boldsymbol q$,
using the same objective $\mathcal J$ and holding $a$, $f$,
$\beta$, and $s$ fixed. These steps are performed whether or
not a curvature correction was accepted. Each of the four
Adam steps evaluates the loss and gradient on the complete
data set at the current parameter values.

Cycles $20,30,\ldots,510$ end with the exponent--scale update
described next, with $\boldsymbol\theta_a$, $\boldsymbol q$,
and $\boldsymbol\eta$ held fixed. These fifty updates,
together with those after cycles 5 and 10, give 52
exponent--scale updates in total. M1 does not update $\beta$
by Adam. The reported estimate is the final iterate at the
end of cycle 510, including its last exponent--scale update.
Appendix~\ref{app:optimizer} specifies the remaining Adam
settings and the handling of invalid numerical proposals.

\subsubsection{Comparing exponents after adjusting the diffusion scale}
\label{sec:m-profile}

Changing $\beta$ changes both the variances and correlations
of the driving increments. We compare candidate exponents
after allowing the global diffusion scale to adjust to each
one. During this operation, $\boldsymbol\theta_a$,
$\boldsymbol q$, and $\boldsymbol\eta$ remain fixed.
For a candidate exponent $\beta$, we multiply the current
diffusion curve by $\gamma>0$ and minimize the Gaussian
negative log likelihood $\mathcal L$ over $\gamma$.

First, we calculate the reconstructed increment vectors
$\boldsymbol r_p$ for all $P$ paths using the current drift
and diffusion, including the current multiplier $s$.
These vectors remain fixed throughout the search.
For a candidate exponent with
$K_\beta=K_{\beta,\boldsymbol\eta}\succ0$, define
\[
 Q(\beta)=
 \sum_{p=1}^{P}
 \boldsymbol r_p^{\mathsf T}K_\beta^{-1}\boldsymbol r_p.
\]
Thus every candidate is evaluated using all $N=Pn$
reconstructed increments.

Multiplying diffusion by $\gamma$ divides each reconstructed
increment by $\gamma$ and adds $\log\gamma$ to the
per-transition log-Jacobian term. Consequently, up to terms
independent of both $\beta$ and $\gamma$, the Gaussian loss is
\[
 \frac{\log\det K_\beta}{2n}
 +\log\gamma
 +\frac{Q(\beta)}{2N\gamma^2}.
\]
For fixed $\beta$ with $Q(\beta)>0$, its minimizer over
$\gamma>0$ satisfies
\begin{equation}
 \widehat\gamma^2(\beta)=\frac{Q(\beta)}{N}.
 \label{eq:m-scale-profile}
\end{equation}
Substitution gives the profiled negative log likelihood
\begin{equation}
 \mathcal P(\beta)=
 \frac12\left\{
 \frac{\log\det K_\beta}{n}
 +\log\frac{Q(\beta)}{N}+1
 \right\}.
 \label{eq:m-beta-profile}
\end{equation}
Only $\beta$ requires a numerical search: for each candidate,
the associated diffusion multiplier is obtained analytically
from Eq.~\eqref{eq:m-scale-profile}.

We first evaluate $\mathcal P$ at the fixed grid points and
the current exponent listed in Eq.~\eqref{eq:app-profile-grid}.
We order the candidate exponents increasingly, identify the
candidate with the smallest profile loss, and refine the search
in the interval between its two immediate neighbors. If the best initial
candidate is an endpoint, the interval runs from that candidate
to its single neighbor. Within this interval, we seek a lower
profile loss at intermediate values of $\beta$, calculating
the optimal diffusion multiplier at each value tested.
Thus the search can return an exponent outside the initial
candidate set; it is a regional numerical minimization,
not a guaranteed global minimization over the full admissible
interval.

Let $\beta_{\rm prop}$ denote the exponent with the lowest
profile loss among all values evaluated, including the initial
candidates and those tested in the refinement interval.
Minimizing this loss is equivalent to maximizing the likelihood
after optimizing the diffusion scale. The proposed pair is
\[
 \beta_{\rm new}=\beta_{\rm prop},
 \qquad
 s_{\rm new}=s\,\widehat\gamma(\beta_{\rm prop}).
\]
If the pair passes the acceptance checks, both values are
updated; otherwise, fitting stops without applying the proposed
pair. Appendix~\ref{app:profile-search} specifies the numerical
search and acceptance checks. The curvature
penalty in $\mathcal J$ remains unchanged throughout this
operation because $\boldsymbol q$ is fixed.

\subsubsection{Testing a local reduction in curvature}
\label{sec:m-curvature}

The local corrections apply the same smoothing principle to the fitted
drift and diffusion curves, using the bounded updates specified below
to limit how much they can alter the fit obtained by Adam. They act on $a$ and $\sigma$
themselves and are separate from the fixed penalty on the
diffusion preactivation $g_{\boldsymbol q}$. No additional
curvature term is added to $\mathcal J$.

We first specify how curvature and curve displacement are
measured. Both calculations use the $J=129$ equally spaced
standardized states $z_k$ from
Eq.~\eqref{eq:m-penalty-grid}, with spacing
$\delta z=(z_+-z_-)/(J-1)$.
Let
\[
 \overline{\sigma}_z=
 \frac1J\sum_{k=1}^{J}\sigma(\bar x+s_xz_k)
\]
be the arithmetic mean of diffusion on this grid.
Define the dimensionless curves
\begin{equation}
 \begin{aligned}
 \widetilde a(z)
 &=\frac{a(\bar x+s_xz)}{s_x/T},\\
 \widetilde\sigma(z)
 &=\frac{\sigma(\bar x+s_xz)}{\overline{\sigma}_z}.
 \end{aligned}
 \label{eq:m-dimensionless-curves}
\end{equation}
The drift is expressed in units of state change over the
observation horizon, and diffusion is divided by its grid
mean. This mean is recomputed for each candidate diffusion
curve. In particular, the global multiplier $s$ cancels
from $\widetilde\sigma$.

For $b\in\{a,\sigma\}$, let $\widetilde b$ denote the
corresponding dimensionless curve. We measure its discrete
curvature by
\begin{equation}
 \begin{aligned}
 R_b
 &=\frac1{J-2}\sum_{k=2}^{J-1}
 \left[
 \frac{
 \widetilde b(z_{k-1})-2\widetilde b(z_k)
 +\widetilde b(z_{k+1})
 }{(\delta z)^2}
 \right]^2.
 \end{aligned}
 \label{eq:app-local-roughness}
\end{equation}
The expression inside the square brackets is the difference
between two adjacent secant slopes, divided by $\delta z$.
Its square measures changes of slope without cancellation
between positive and negative differences. In particular,
$R_b=0$ when the grid values lie on an affine function.
A reduction in curvature below means a reduction in this
finite-grid quantity.

For any proposed change, write
$\widetilde b_{\rm before}$ and $\widetilde b_{\rm after}$
for the dimensionless curves before and after that change.
Their displacement is measured by the root mean square
(RMS) difference
\[
 \left\{
 \frac1J\sum_{k=1}^{J}
 \left[
 \widetilde b_{\rm after}(z_k)
 -\widetilde b_{\rm before}(z_k)
 \right]^2
 \right\}^{1/2}.
\]
We use this same definition to measure the curve displacement
produced by the preceding joint Adam step.

Corrections are tested only when that joint step decreases
$\mathcal J$. Let $\mathcal J_{\rm old}$ and
$\mathcal J_{\rm Adam}$ be the objective values immediately
before and after the joint step, and define
\[
 \Delta\mathcal J=
 \mathcal J_{\rm old}-\mathcal J_{\rm Adam}>0.
\]
Starting from the post-Adam parameters, we test the drift
correction first and the diffusion correction second.
Only $\boldsymbol\theta_a$ changes during the drift test,
and only $\boldsymbol q$ changes during the diffusion test;
$\boldsymbol\eta$, $\beta$, and $s$ remain fixed.

For each function, the proposed direction is obtained from
the negative gradient of $R_b/2$, scaled componentwise using
Adam's second moments. This direction is intended to reduce
$R_b$; the actual reduction is checked for every candidate.
We test up to thirteen parameter displacements in this
direction, halving their size successively.
Appendix~\ref{app:implementation} gives the direction and
the initial displacement.

Set $\rho_a=0.05$ and $\rho_\sigma=0.01$.
A candidate correction is accepted only if all four
conditions below hold:

\begin{enumerate}

\item \emph{Parameter displacement.}
The Euclidean norm of the parameter change introduced by
the correction must be at most $\rho_b$ times the norm of
the change produced by the preceding joint Adam step for
the same function.

\item \emph{Curve displacement.}
The RMS displacement of the dimensionless curve introduced
by the correction must be at most $\rho_b$ times its RMS
displacement in the preceding joint Adam step. This condition
bounds the change in the fitted curve separately from the
change in its parameters.

\item \emph{Curvature reduction.}
Let $R_b^{\rm before}$ and $R_b^{\rm after}$ be the discrete
curvature values immediately before and after the candidate
correction. We require
\begin{equation}
 (1-\rho_b)R_b^{\rm before}
 \leq R_b^{\rm after}
 <R_b^{\rm before}.
 \label{eq:m-roughness-bound}
\end{equation}
Thus curvature must decrease, but by no more than $5\%$
of its value before the drift correction or $1\%$ of its
value before the diffusion correction. These percentages
are upper bounds, not required reductions.

\item \emph{Objective change.}
Let $\mathcal J_{\rm current}$ and
$\mathcal J_{\rm candidate}$ be the full objective values
immediately before and after the candidate correction.
We require
\[
 \mathcal J_{\rm candidate}-\mathcal J_{\rm current}
 \leq \rho_b\Delta\mathcal J.
\]
The permitted increase is therefore a fraction of the
decrease $\Delta\mathcal J$ obtained by the preceding joint
Adam step, not a fraction of the objective value itself.
A correction that decreases $\mathcal J$ also satisfies
this condition, but must still satisfy the other three.
When diffusion is tested, $\mathcal J_{\rm current}$
includes any accepted drift correction.

\end{enumerate}

We accept the first candidate satisfying all four conditions.
If none does, the selected function keeps its post-Adam
parameters. Both objective allowances are calculated from
the same $\Delta\mathcal J$. Hence, if
$\mathcal J_{\rm corr}$ denotes the objective after both
correction tests,
\[
 \mathcal J_{\rm corr}
 \leq \mathcal J_{\rm Adam}+0.06\Delta\mathcal J
 =\mathcal J_{\rm old}-0.94\Delta\mathcal J.
\]
The corrections therefore preserve at least $94\%$ of the
decrease obtained by the joint Adam step.

Accepted corrections change the function parameters without
changing Adam's stored moments. The outcome of each drift
and diffusion correction is recorded. The objective and
curvature bounds apply immediately after these corrections;
the three subsequent Adam steps on $\boldsymbol q$ can
change both quantities again.

\subsection{Methods 2 and 3: neural likelihood estimation}
\label{sec:m-comparators}

Methods 2 and 3 use the same Gaussian trajectory likelihood
and temporal-weight normalization as M1. They represent
each unknown function by a neural network and use Adam
to update the network parameters and a transformed
covariance exponent. Their likelihood-based updates
minimize $\mathcal L$, without the curvature penalty
included in M1's objective $\mathcal J$. Neither method
uses a separate diffusion-scale multiplier, an
exponent--scale profile, or local curvature corrections.

Both methods use a drift network with four hidden layers
of 32 ELU units. The diffusion and temporal-weight
networks each have three hidden layers of 32 ELU units.
All three networks have an affine output layer.
The drift and diffusion networks receive the standardized
state input $\tanh z(x)$; the temporal network receives
$\tanh(2t/T-1)$. The drift is the untransformed output
of its network. For diffusion, we apply softplus to
the network output and add $10^{-8}$, so the fitted
coefficient is positive. Its shape and amplitude are
both determined by the diffusion-network parameters.

For the temporal weight, we apply softplus and divide
the resulting positive output by its mean at the
$L=24$ integration-cell midpoints, using the normalization
in Eq.~\eqref{eq:m-weight-net}. The denominator is
recomputed whenever the temporal network changes.
The resulting midpoint values determine the full
increment covariance through
Eq.~\eqref{eq:m-product-K}.

The covariance exponent is parameterized by a logistic
transformation of an unconstrained real scalar.
Whenever the exponent is updated, Adam acts on this
scalar, and the transformation gives the new value
of $\beta$. Appendix~\ref{app:optimizer} specifies
the transformation and optimizer settings.

Every fit starts from the same constant drift,
diffusion, and temporal-weight functions as M1,
with $\beta=1$ and training seed 12. Each gradient
evaluation uses the complete data set, as described
in Sec.~\ref{sec:m-data-evaluation}.
The following subsections specify the losses and
active parameters at each stage. With the
representations and initialization above, M2 and M3
differ only in their first ten training cycles.

\subsubsection{Method 2: fit all components from the first cycle}
\label{sec:m2-training}

M2 performs one joint Adam step per cycle on the drift,
diffusion, and temporal-network parameters and on the
unconstrained scalar that determines $\beta$ through
the logistic transformation. Each step evaluates
$\mathcal L$ and its gradient using all observed paths.
All four unknown components are therefore estimated
jointly from the first cycle. This update is repeated
for 510 cycles, and the parameters obtained after
the last cycle define the reported estimate.

This procedure adapts the joint Euler--Maruyama
likelihood approach of
Dietrich \textit{et al.}\cite{Dietrich2023}, who trained
drift and diffusion networks using Gaussian transition
densities for Brownian-driven equations. In that
setting, the conditional mean and variance of the
state increment are $h_i a(X_{i,p})$ and
$h_i\sigma(X_{i,p})^2$, respectively.
Setting $K=\operatorname{diag}(h_0,\ldots,h_{n-1})$
in Eq.~\eqref{eq:weighted-likelihood} gives the
corresponding Gaussian transition criterion, up to
an additive constant and a positive multiplicative
factor.

Our adaptation uses the full increment covariance
$K(\beta,f)$ to account for temporal dependence
within each path and estimates $f$ and $\beta$
jointly with the drift and diffusion.
The network architectures, initialization, and
510-cycle training schedule are those specified
for the present comparison.

\subsubsection{Method 3: initialize drift and noise separately}
\label{sec:m3-training}

M3 begins by fitting the drift and noise components
in separate stages. During cycles 1--5, each cycle
takes one Adam step on the drift-network parameters,
using $\mathcal L_a$ in Eq.~\eqref{eq:m-warmup}.
The diffusion network, temporal network, and exponent
retain their initial values.

During cycles 6--10, the initialized drift is held
fixed. Each cycle takes one joint Adam step on the
diffusion-network parameters, temporal-network
parameters, and unconstrained scalar that determines
$\beta$, using the Gaussian loss $\mathcal L$.
The first exponent update occurs in cycle 6.
No exponent--scale profile is performed between
these stages.

During cycles 11--510, each cycle takes one joint
Adam step on all three networks and the transformed
exponent, using $\mathcal L$, as in M2.
The drift can therefore change again while the noise
components continue to be estimated. The reported
estimate is the fit after cycle 510.

A stage transition changes only which parameters are
active and which loss is used. Parameter values,
Adam's first and second moments, and the update
counter for each parameter are retained.
While a parameter is inactive, its value, moments,
and counter remain unchanged.

The drift-then-noise initialization adapts the
two-stage procedure of
Fang \textit{et al.}\cite{Fang2022}.
For symmetric $\alpha$-stable L\'evy forcing with
a fixed stability index $0<\alpha<2$, $\alpha\ne1$,
they first fitted the drift by minimizing squared
Euler prediction errors. They then held that drift
fixed and fitted the diffusion coefficient through
the stable transition likelihood.
Here, the second stage uses the correlated Gaussian
trajectory likelihood and estimates the temporal
weight and covariance exponent together with
diffusion. The five-cycle duration of each initial
stage and the subsequent 500 joint updates are
choices of the present adaptation.

Although all three methods run for 510 cycles,
a cycle contains different operations in each method.
M2 and M3 each perform 510 Adam steps.
M1 performs five initial drift steps and four Adam
steps in each of the remaining 505 cycles, giving
$5+4\times505=2025$ Adam steps, in addition to its
curvature tests and 52 exponent--scale updates.
The numerical comparison evaluates these complete
procedures with the training schedules specified
above.

\section{A worked nonlinear experiment}\label{sec:example}
\subsection{Generating coefficients and observed trajectories}
We begin with a model in which the drift pulls the state toward zero, the diffusion varies with the state, and the temporal weight increases with time. The generating coefficients are

\begin{align}
 a_*(x)&=-\tanh x,\qquad \beta_*=1.5,\nonumber\\
 \sigma_*(x)&=0.5+0.2\sin x,\nonumber\\
 f_*(t)&=\frac23+\frac{2t}{15},\qquad 0\leq t\leq5.
 \label{eq:example-functions}
\end{align}

The subscript $*$ denotes a generating quantity. The diffusion coefficient lies between $0.3$ and $0.7$, while $f_*$ increases from $2/3$ to $4/3$ and has mean one on $[0,5]$. Both state functions are smooth and bounded, with bounded derivatives, and the temporal weight is positive.

To observe the dynamics on both sides of zero, we use the five initial states $-5,-2.5,0,2.5,5$ and simulate five independent paths from each. Recording at $t_i=0.005i$, $i=0,\ldots,1000$, gives $P=25$ trajectories on $[0,5]$, comprising $25\,000$ transitions and $25\,025$ states. The interval $[-5,5]$ specifies the starting points; it does not constrain the subsequent motion.

\begin{figure*}[tbp]
\centering\manuscriptfigure[width=.95\textwidth]{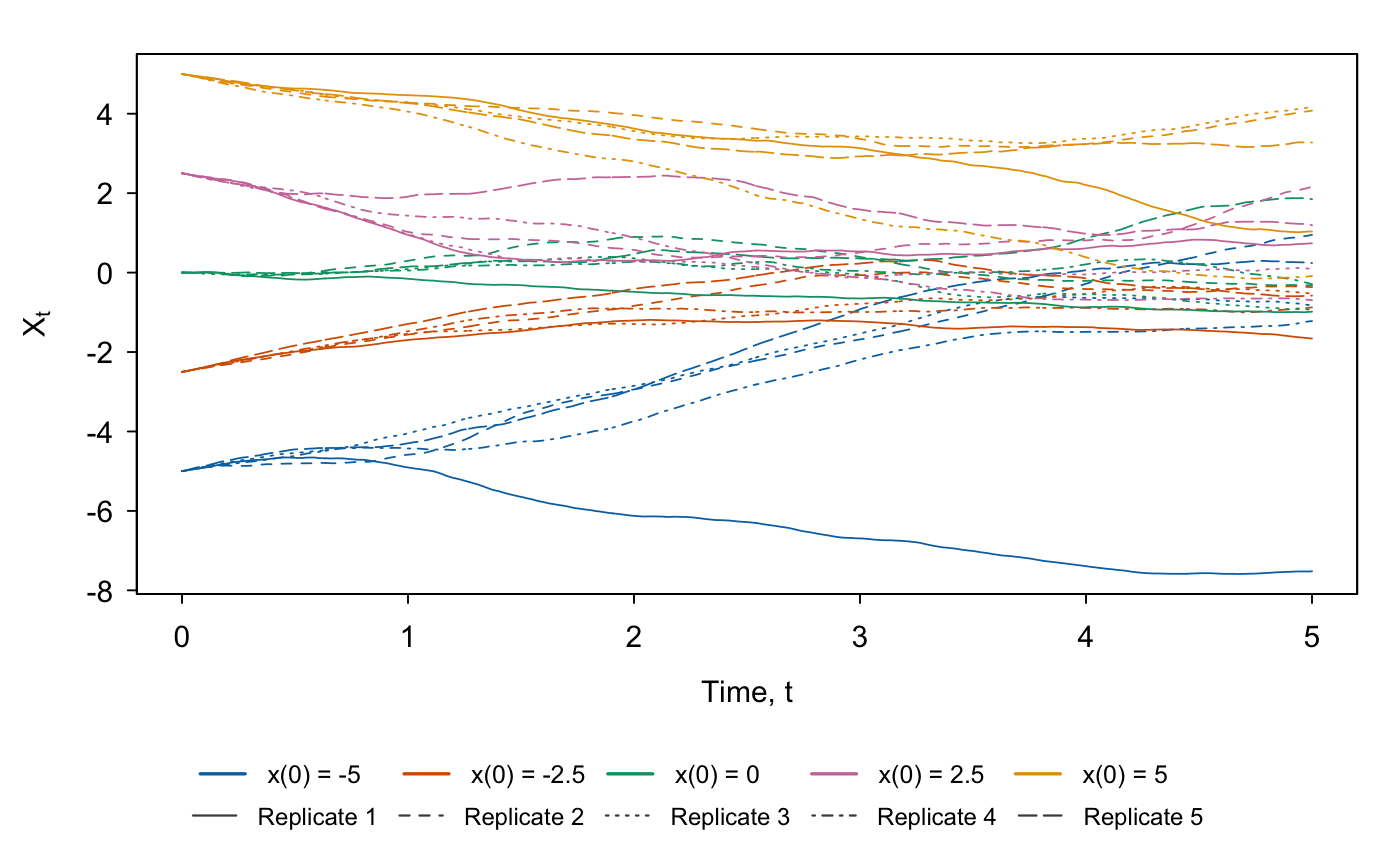}
\caption{Observed trajectories for the coefficients in Eq.~\eqref{eq:example-functions} (case 1). Colors distinguish the five initial states; line styles distinguish the five independent paths from each initial state. Every path contains 1001 observations.}
\label{fig:trajectories}
\end{figure*}

Simulation follows the Euler recursion. We first replace $f_*$ by its average on each of 32 equal time cells and calculate the resulting increment covariance analytically. For each path, we multiply its lower-triangular Cholesky factor by a fresh vector of 1000 independent standard-normal variables. The resulting correlated increments drive the Euler updates. With seed 923, this gives the trajectories in Fig.~\ref{fig:trajectories}, whose states span approximately $[-7.589,5]$. Each method receives the same path identifiers, times, and states. The estimators receive only these observations; the generating quantities are retained for evaluating the fitted functions.

\subsection{The fit and its numerical evaluation}
\begin{figure*}[tbp]
\centering\manuscriptfigure[width=\textwidth]{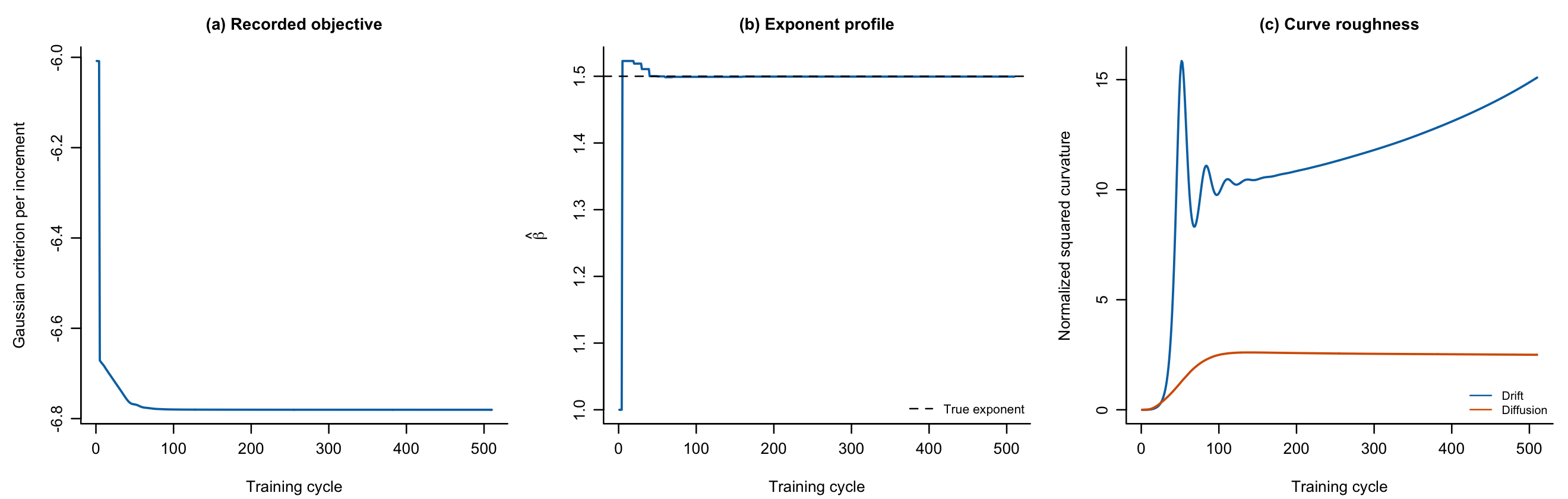}
\caption{Method-1 training in case 1: (a) Gaussian negative log likelihood per transition, excluding $\log(2\pi)/2$; (b) profiled exponent, with the generating value shown for reference; and (c) squared curvature of the dimensionless state curves defined in Eq.~\eqref{eq:m-dimensionless-curves}. Each entry is recorded at the end of a complete cycle, after the additional diffusion updates and any scheduled exponent--scale profile.}
\label{fig:training}
\end{figure*}

We fit M1 with seed 12 using the 510-cycle schedule in Sec.~\ref{sec:m1-training}. Figure~\ref{fig:training} shows the resulting training history, including the initial drift fit. The fitted covariance uses 24 midpoint cells and all observation times. After the initial drift fit, the exponent--scale update at cycle 5 gives $\widehat\beta=1.52290413$. The later updates bring this estimate to $1.49952243$, while the Gaussian negative log likelihood per transition decreases from $-6.00813723$ to $-6.78041220$.

The drift starts from a constant and must develop curvature to represent $-\tanh x$. This is consistent with the histories in Fig.~\ref{fig:training}(c): each accepted local correction reduces the curvature of the curve produced by the preceding joint update, but the Adam steps can introduce curvature. Consequently, curvature need not decrease from one completed cycle to the next.

For the drift and diffusion coefficient, we evaluate accuracy at the states actually observed. If $\widehat g$ estimates $g_*$, with $g\in\{a,\sigma\}$, its mean squared error (MSE) is

\begin{equation}
 \operatorname{MSE}_g=\frac1{P(n+1)}\sum_{p=1}^{P}\sum_{i=0}^{n}
 [\widehat g(X_{i,p})-g_*(X_{i,p})]^2.
 \label{eq:example-state-errors}
\end{equation}

This average includes all $25\,025$ states, including the initial and terminal states and repeated visits. Regions visited more often therefore receive more weight, as prescribed by the empirical distribution of the observations.

The temporal weight is evaluated over the observation interval. We calculate its integrated squared error (ISE) and divide by the interval length to obtain a temporal MSE:

\begin{align}
 \operatorname{ISE}_f&=\int_0^T[\widehat f(t)-f_*(t)]^2\,dt,
 \label{eq:example-weight-errors}\\
 \operatorname{MSE}_f&=\operatorname{ISE}_f/T.
 \label{eq:mse-f-all}
\end{align}

Thus $\operatorname{ISE}_f$ is the squared $L^2(0,T)$ error. Adaptive quadrature integrates the squared difference between the continuous fitted and generating functions, with numerical settings specified in the accompanying code. The piecewise constant weights used to assemble the covariance do not enter this error calculation. The exponent is evaluated by its squared error, $(\widehat\beta-\beta_*)^2$. We use these same error definitions for all fits.

\begin{figure*}[tbp]
\centering\manuscriptfigure[width=\textwidth]{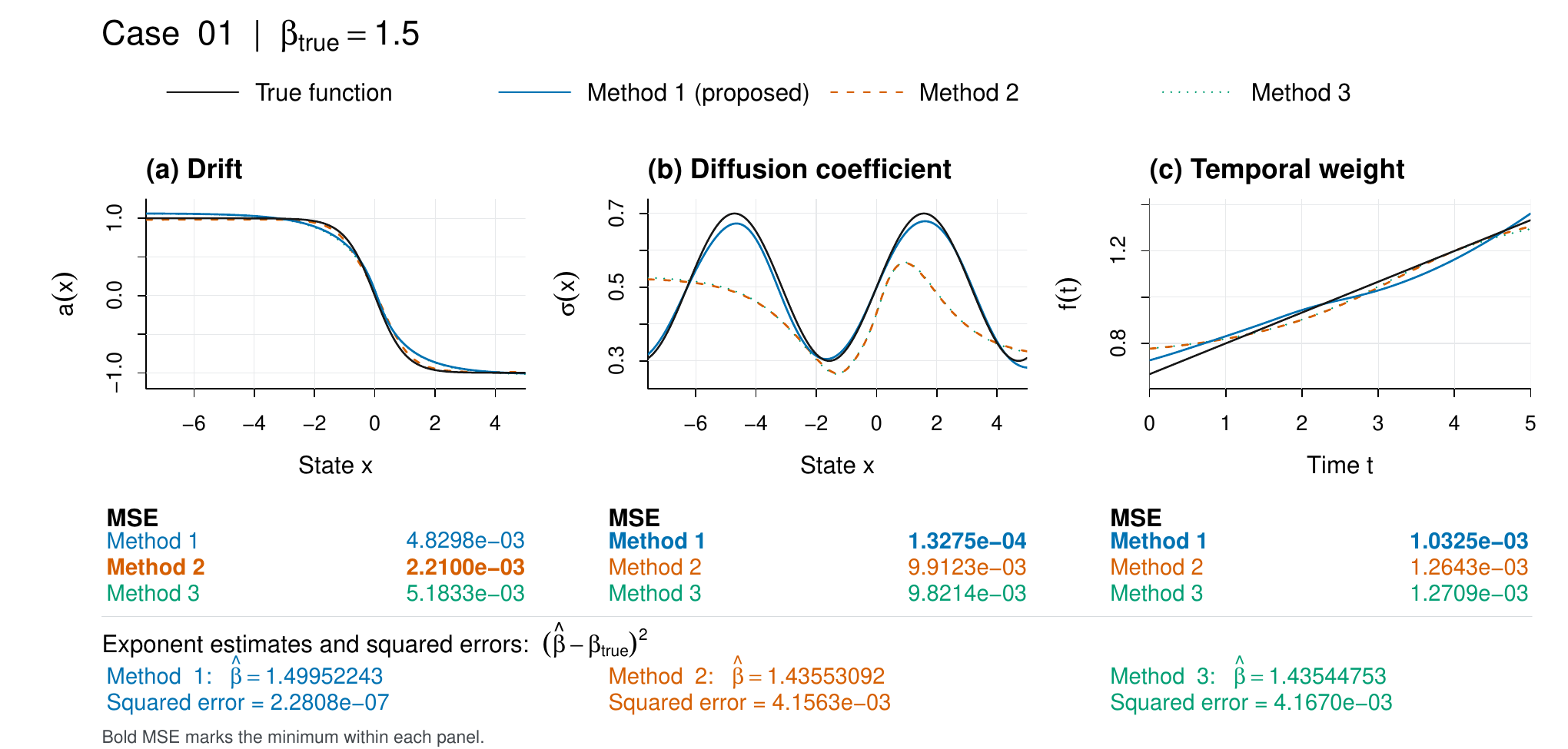}
\caption{Drift, diffusion coefficient, and temporal weight recovered in case 1. Each panel shows the generating function and the three fitted curves, together with their MSEs. The footer gives the exponent estimates and squared errors. State-function MSEs use all recorded states, and the temporal MSE is the time average of the squared discrepancy between the continuous curves.}
\label{fig:coefficients}
\end{figure*}

In this example, M1 has the smallest errors for the diffusion, temporal weight, and exponent (Fig.~\ref{fig:coefficients}). Its diffusion MSE is $1.327\times10^{-4}$ and its temporal MSE is $1.032\times10^{-3}$, corresponding to ISE $5.162\times10^{-3}$; the squared exponent error is $2.281\times10^{-7}$. The drift gives a different result: M2 has MSE $2.210\times10^{-3}$, compared with $4.830\times10^{-3}$ for M1. This motivates examining each component separately in the broader comparison.

\section{Comparison across twenty coefficient settings}\label{sec:comparisons}
\subsection{Experimental design}
We keep the observation design of the example and choose the generating
coefficients from a catalogue of five test functions for each component.
Table~\ref{tab:test-functions} lists the fifteen functions, denoted by
$a_j$, $\sigma_j$, and $f_j$, with $j=1,\ldots,5$. The state functions
are defined for $x\in\mathbb R$, and the temporal weights for
$t\in[0,5]$. Each weight satisfies $5^{-1}\int_0^5 f_j(t)\,dt=1$.
For the fifth weight, this normalization uses the mean of the
unnormalized bell-shaped function,
\begin{equation}
 c_5=\frac15\int_0^5\left[1+2e^{-((u-2.5)/0.7)^2}\right]\,du
 \simeq1.49628686.
 \label{eq:bell-normalizer}
\end{equation}
Thus $c_5$ is a fixed constant in the generating function $f_5$.

\begin{table*}[tbp]
\caption{Five test functions for each component of the generating model.
The drift and diffusion are defined on $\mathbb R$ and the temporal
weights on $[0,5]$. The constant $c_5$ is defined in
Eq.~\eqref{eq:bell-normalizer}. Each component is selected independently
by its own index.}
\label{tab:test-functions}
\centering
\begingroup\small
\renewcommand{\arraystretch}{1.65}
\begin{tabular*}{\textwidth}{@{\extracolsep{\fill}}c c c c@{}}
\toprule
Index $j$ & Drift $a_j(x)$ & Diffusion $\sigma_j(x)$ & Temporal weight $f_j(t)$ \\
\midrule
1 & $-\dfrac{x}{1+x^2}$ & $0.5+0.2\tanh^2 x$ & $1$ \\
2 & $-\tanh x$ & $0.4$ & $\dfrac23(1+t/5)$ \\
3 & $0.5\sin x-0.25\tanh x$ & $0.35+\dfrac{0.15}{1+x^2}$ & $1+0.5\sin(2\pi t/5)$ \\
4 & $\dfrac{1-x}{1+(x-1)^2}$ & $0.5+0.2\sin x$ & $\dfrac{e^{-t/5}}{1-e^{-1}}$ \\
5 & $-0.8\tanh(x-1.5)+0.2\sin x$ & $0.3+0.4e^{-(x-1.5)^2}$ & $\dfrac{1+2e^{-((t-2.5)/0.7)^2}}{c_5}$ \\
\bottomrule
\end{tabular*}
\endgroup
\end{table*}

We fix the five exponent values $\beta_*=1.5,1.8,1,0.5,0.2$.
For each value, we draw four index triples $(j_a,j_\sigma,j_f)$.
Every index is drawn independently and uniformly from
$\{1,\ldots,5\}$, with replacement across the draws. The first index
selects the drift, the second the diffusion, and the third the temporal
weight, giving
\[
 (a_*,\sigma_*,f_*)=(a_{j_a},\sigma_{j_\sigma},f_{j_f}).
\]
Repeated function choices are therefore allowed. Four triples for each
of the five exponents give twenty settings. The selected triples and
their generating functions are reported in
Tables~\ref{tab:cases-1p5}--\ref{tab:cases-0p2}. This catalogue specifies
the simulations and the reference functions used to evaluate error;
the estimators fit the unknown functions from the observations without
being given the selected indices or generating functions.

For every setting, we simulate five independent paths from each initial
state in $\{-5,-2.5,0,2.5,5\}$, with 1000 transitions per path over
$[0,5]$. The three methods receive identical saved observations within
a setting, as checked from the observation tables and their checksums.
Each setting is fitted separately: every Adam step uses its $P=25$
complete paths and $N=25\,000$ transitions, with a common
$1000\times1000$ increment covariance for Gaussian loss evaluations.
Each method follows its prescribed 510-cycle schedule, including its
own representations and numbers of updates.

The numerical construction is also unchanged: the generating covariance uses weight averages on 32 cells, the fitted covariance uses 24 midpoint values, and errors are calculated from the continuous temporal curves. At $\beta_*=1$, covariance evaluation uses the logarithmic kernel in Eq.~\eqref{eq:weighted-q}.

We reset the simulation seed to 923 and the training seed to 12 in every setting. Reusing the simulation seed means that different coefficient combinations act on the same underlying standard-normal draws. The 25 driving vectors within each setting remain independent. There is one saved data set per setting and one completed fit per method, yielding the sixty fits below.

\subsection{Accuracy of the estimated components}
The methods differ most consistently in estimation of the noise components. M1 has the smallest squared exponent error in 19 of twenty settings, the smallest diffusion MSE in 16, and the smallest temporal-weight MSE in 14 (Table~\ref{tab:paired-comparators}). Drift estimation is less uniform: M1 has the smallest MSE in ten settings, and M2 and M3 each in five. Together these give M1 59 of the 80 componentwise minima. Temporal ISE is not counted a second time, since $T=5$ throughout and $\operatorname{ISE}_f=5\operatorname{MSE}_f$.

\par\medskip\noindent\begin{minipage}{\columnwidth}
\captionof{table}{Frequency with which each method has the smallest componentwise error across the twenty cases. Each entry shows the percentage above the count. Function columns use MSE and the exponent column uses squared error. Temporal ISE is omitted because it is proportional to temporal MSE; there are no ties.}
\label{tab:paired-comparators}
\centering\small\setlength{\tabcolsep}{2pt}
\begin{tabular*}{\columnwidth}{@{\extracolsep{\fill}}lrrrr@{}}
\toprule
Method & $a$ & $\sigma$ & $f$ & $\beta$ \\
\midrule
M1 & \shortstack{\textbf{50\%}\\\textbf{10/20}} & \shortstack{\textbf{80\%}\\\textbf{16/20}} & \shortstack{\textbf{70\%}\\\textbf{14/20}} & \shortstack{\textbf{95\%}\\\textbf{19/20}} \\
M2 & \shortstack{25\%\\5/20} & \shortstack{15\%\\3/20} & \shortstack{20\%\\4/20} & \shortstack{5\%\\1/20} \\
M3 & \shortstack{25\%\\5/20} & \shortstack{5\%\\1/20} & \shortstack{10\%\\2/20} & \shortstack{0\%\\0/20} \\
\bottomrule
\end{tabular*}
\end{minipage}\par\medskip

For M1, diffusion and temporal-weight MSEs range over $2.071\times10^{-5}$--$2.018\times10^{-3}$ and $2.494\times10^{-4}$--$2.123\times10^{-3}$, with medians $5.548\times10^{-4}$ and $9.754\times10^{-4}$, respectively. Drift MSE ranges from $1.666\times10^{-3}$ to $4.291\times10^{-2}$, with median $1.052\times10^{-2}$. The largest squared exponent error is $2.006\times10^{-5}$. Some exponent errors are at numerical precision because the profile search retains a grid candidate equal to the generating value; they describe the fitted point estimate, not its uncertainty.

\subsection{Differences across covariance regimes}
The drift ranking changes with the generating exponent (Table~\ref{tab:regime-wins}). When $\beta_*>1$, M1 attains the smallest drift MSE in only two of eight settings, although it has the smallest errors for $\sigma$, $f$, and $\beta$ in all eight. When $\beta_*<1$, it attains the smallest drift MSE in six of eight settings and the smallest diffusion and exponent errors in every setting. At $\beta_*=1$, the comparison changes again: M1 attains six of sixteen componentwise minima, and a comparator has the smaller diffusion MSE in all four settings.

\par\medskip\noindent\begin{minipage}{\columnwidth}
\captionof{table}{Method-1 component wins by covariance regime. Function columns use MSE and the exponent column uses squared error; the total counts all four components in each case.}
\label{tab:regime-wins}
\centering\small\setlength{\tabcolsep}{4pt}
\begin{tabular}{lrrrrr}
\toprule
Regime & $a$ & $\sigma$ & $f$ & $\beta$ & Total\\
\midrule
$\beta_*>1$ & 2/8 & 8/8 & 8/8 & 8/8 & 26/32\\
$\beta_*=1$ & 2/4 & 0/4 & 1/4 & 3/4 & 6/16\\
$\beta_*<1$ & 6/8 & 8/8 & 5/8 & 8/8 & 27/32\\
\bottomrule
\end{tabular}
\end{minipage}\par\medskip

We measure M1's excess error relative to the better of the two comparators. For component $g\in\{a,\sigma,f,\beta\}$, define

\begin{equation}
 \Delta_g=\operatorname{MSE}_{g,1}
 -\min\{\operatorname{MSE}_{g,2},\operatorname{MSE}_{g,3}\},
 \label{eq:loss-gap}
\end{equation}

Here $\operatorname{MSE}_{g,j}$ denotes the error of method $j$ for component $g$, using squared exponent error when $g=\beta$. A positive gap is the excess error of M1 over the better comparator. Table~\ref{tab:loss-gaps} summarizes these positive gaps.

The largest drift gap occurs in case 2: M1 has MSE $2.304\times10^{-2}$ and M2 has MSE $9.522\times10^{-3}$, a difference of $1.352\times10^{-2}$. Across the ten settings where M1 does not have the smallest drift MSE, the gaps range from $1.891\times10^{-4}$ to $1.352\times10^{-2}$. Eight are between $10^{-3}$ and $10^{-2}$, one between $10^{-4}$ and $10^{-3}$, and one between $10^{-2}$ and $10^{-1}$. By comparison, all four positive diffusion gaps occur at $\beta_*=1$ and are below $10^{-3}$. The six temporal-weight gaps range from $5.499\times10^{-5}$ to $1.205\times10^{-3}$. The sole positive exponent gap is $1.840\times10^{-5}$ in case 12.

\par\medskip\noindent\begin{minipage}{\columnwidth}
\captionof{table}{Excess error of method 1 in cases where a comparator has smaller error, measured by Eq.~\eqref{eq:loss-gap}. Only positive gaps are included.}
\label{tab:loss-gaps}
\centering\small\setlength{\tabcolsep}{3pt}
\begin{tabular}{lrrr}
\toprule
Component & Cases & Minimum gap & Maximum gap\\
\midrule
$a$ & 10 & $1.891\times10^{-4}$ & $1.352\times10^{-2}$\\
$\sigma$ & 4 & $3.132\times10^{-4}$ & $9.996\times10^{-4}$\\
$f$ & 6 & $5.499\times10^{-5}$ & $1.205\times10^{-3}$\\
$\beta$ & 1 & $1.840\times10^{-5}$ & $1.840\times10^{-5}$\\
\bottomrule
\end{tabular}
\end{minipage}\par\medskip

\begin{figure*}[tbp]
\centering\manuscriptfigure[width=\textwidth]{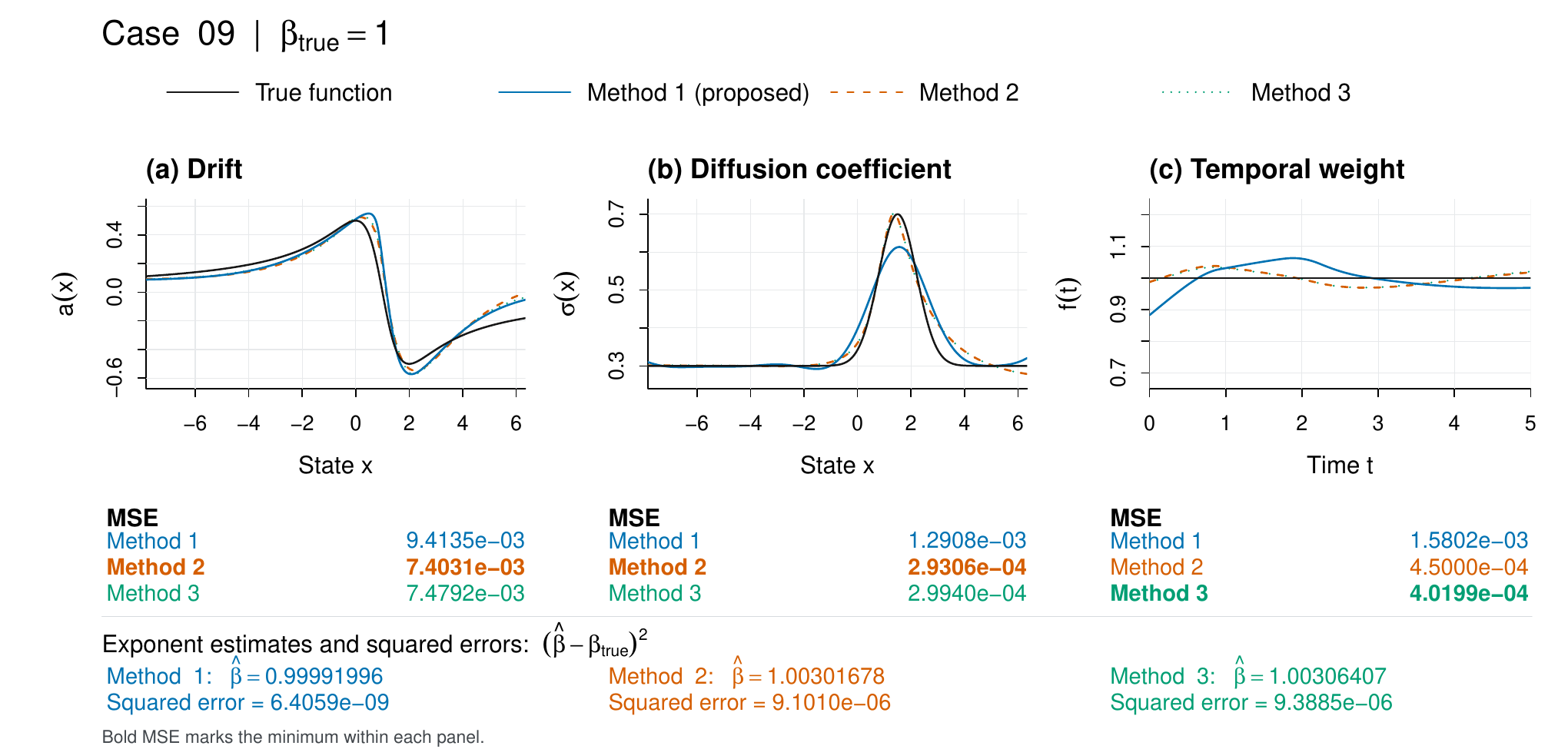}
\caption{Case 9, $\beta_*=1$. Each panel compares the generating function with the three estimates and reports their MSEs; the footer gives the exponent estimates and squared errors. The comparators estimate all three functions more accurately, while method 1 has the smallest exponent error.}
\label{fig:case-nine}
\end{figure*}

\begin{figure*}[tbp]
\centering\manuscriptfigure[width=\textwidth]{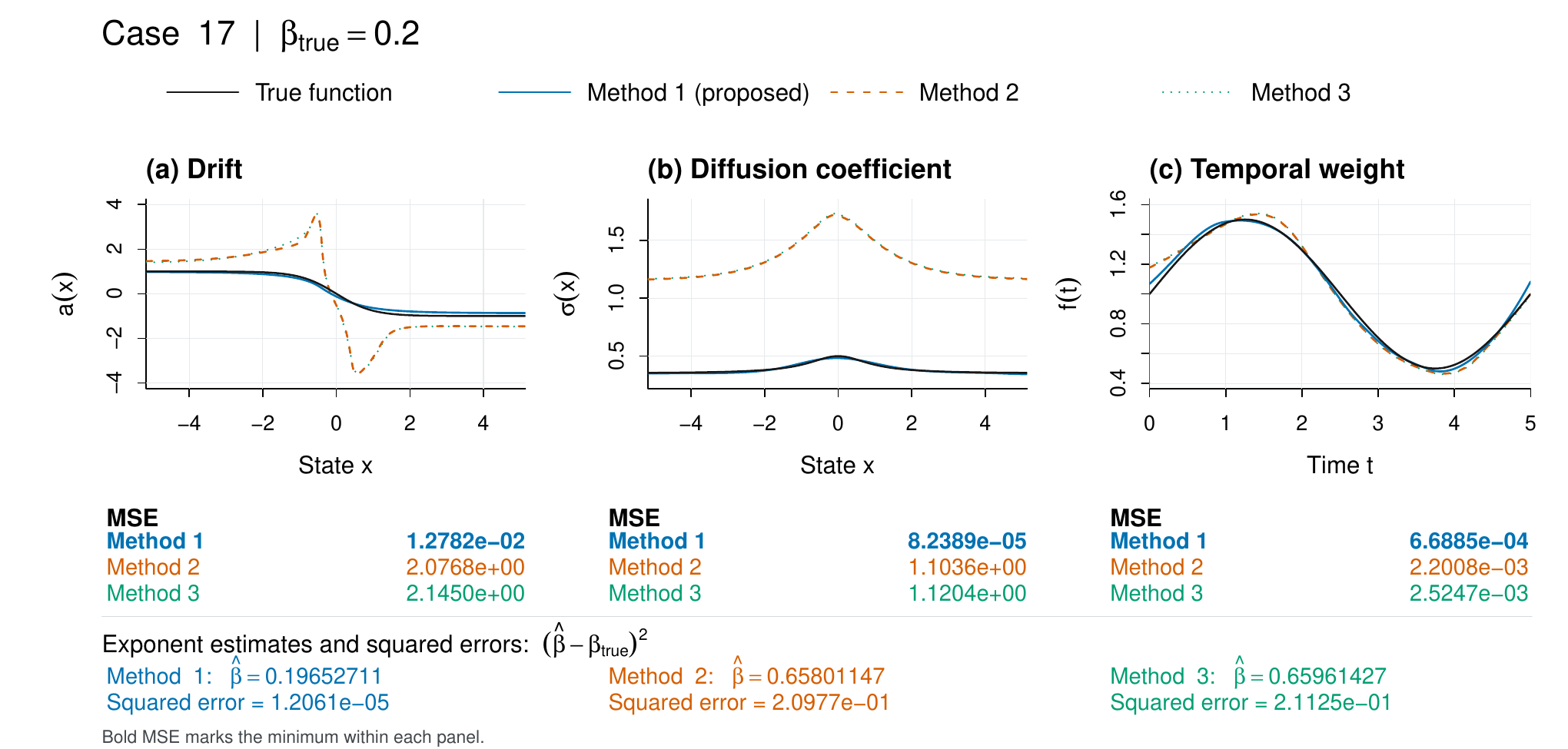}
\caption{Case 17, $\beta_*=0.2$. Panels show the generating and fitted functions with their MSEs; the footer gives the exponent estimates and squared errors. Method 1 has the smallest error for all four components.}
\label{fig:case-seventeen}
\end{figure*}

Figures~\ref{fig:case-nine} and \ref{fig:case-seventeen} compare the fitted functions in two contrasting settings. In case 9, at $\beta_*=1$, both M2 and M3 recover all three functions more accurately than M1, while M1 gives the smallest exponent error (Fig.~\ref{fig:case-nine}). In case 17, at $\beta_*=0.2$, M1 gives the smallest error for every component (Fig.~\ref{fig:case-seventeen}). Its drift MSE is $1.278\times10^{-2}$, compared with $2.077$ for M2 and $2.145$ for M3. The corresponding exponent estimates are $0.19653$, $0.65801$, and $0.65961$.

Tables~\ref{tab:cases-1p5}--\ref{tab:cases-0p2} report every fit alongside its generating functions. In panel (a), the parenthesized indices are $(j_a,j_\sigma,j_f)$ and refer to the function catalogue in Table~\ref{tab:test-functions}. Panel (b) gives the estimated exponent and all errors, with the smallest error in each component shown in boldface.

\onecolumngrid
\noindent\begin{minipage}{\textwidth}
\centering
\captionof{table}{Generating functions and estimation errors at $\beta_*=1.5$. Methods share the observations within each case; boldface identifies the smallest error for that component and case. $\operatorname{ISE}_f=\|\widehat f-f_*\|_{L^2(0,5)}^2=5\operatorname{MSE}_f$.}
\label{tab:cases-1p5}
\begingroup\small
\renewcommand{\arraystretch}{1.30}
\begin{tabular*}{\textwidth}{@{\extracolsep{\fill}}lccc@{}}
\toprule
\multicolumn{4}{l}{\textbf{(a) Generating functions; $c_5$ is defined in Eq.~\eqref{eq:bell-normalizer}}}\\
Case (indices) & $a_*(x)$ & $\sigma_*(x)$ & $f_*(t)$ \\
\midrule
1 (2,4,2) & $-\tanh x$ & $0.5+0.2\sin x$ & $\frac23(1+t/5)$ \\
2 (3,3,3) & $0.5\sin x-0.25\tanh x$ & $0.35+\frac{0.15}{1+x^2}$ & $1+0.5\sin(2\pi t/5)$ \\
3 (4,2,4) & $\frac{1-x}{1+(x-1)^2}$ & $0.4$ & $\frac{e^{-t/5}}{1-e^{-1}}$ \\
4 (1,5,3) & $-\frac{x}{1+x^2}$ & $0.3+0.4e^{-(x-1.5)^2}$ & $1+0.5\sin(2\pi t/5)$ \\
\bottomrule\end{tabular*}
\par\medskip
\renewcommand{\arraystretch}{1.06}
\setlength{\tabcolsep}{3pt}
\begin{tabular*}{\textwidth}{@{\extracolsep{\fill}}llrrrrrr@{}}
\toprule
\multicolumn{8}{l}{\textbf{(b) Squared-error measures}}\\
Case & Method & $\widehat\beta$ & $(\widehat\beta-\beta_*)^2$ & $\operatorname{MSE}_a$ & $\operatorname{MSE}_{\sigma}$ & $\operatorname{MSE}_f$ & $\operatorname{ISE}_f$ \\
\midrule
1 & M1 & 1.499522 & $\boldsymbol{2.281\!\times\!10^{-7}}$ & $4.830\!\times\!10^{-3}$ & $\boldsymbol{1.327\!\times\!10^{-4}}$ & $\boldsymbol{1.032\!\times\!10^{-3}}$ & $\boldsymbol{5.162\!\times\!10^{-3}}$ \\
1 & M2 & 1.435531 & $4.156\!\times\!10^{-3}$ & $\boldsymbol{2.210\!\times\!10^{-3}}$ & $9.912\!\times\!10^{-3}$ & $1.264\!\times\!10^{-3}$ & $6.321\!\times\!10^{-3}$ \\
1 & M3 & 1.435448 & $4.167\!\times\!10^{-3}$ & $5.183\!\times\!10^{-3}$ & $9.821\!\times\!10^{-3}$ & $1.271\!\times\!10^{-3}$ & $6.355\!\times\!10^{-3}$ \\
\addlinespace[3pt]
2 & M1 & 1.502506 & $\boldsymbol{6.280\!\times\!10^{-6}}$ & $2.304\!\times\!10^{-2}$ & $\boldsymbol{5.862\!\times\!10^{-5}}$ & $\boldsymbol{6.737\!\times\!10^{-4}}$ & $\boldsymbol{3.368\!\times\!10^{-3}}$ \\
2 & M2 & 1.410343 & $8.038\!\times\!10^{-3}$ & $\boldsymbol{9.522\!\times\!10^{-3}}$ & $6.162\!\times\!10^{-3}$ & $2.809\!\times\!10^{-2}$ & $1.404\!\times\!10^{-1}$ \\
2 & M3 & 1.410654 & $7.983\!\times\!10^{-3}$ & $2.181\!\times\!10^{-2}$ & $6.112\!\times\!10^{-3}$ & $2.966\!\times\!10^{-2}$ & $1.483\!\times\!10^{-1}$ \\
\addlinespace[3pt]
3 & M1 & 1.500505 & $\boldsymbol{2.554\!\times\!10^{-7}}$ & $5.008\!\times\!10^{-3}$ & $\boldsymbol{2.071\!\times\!10^{-5}}$ & $\boldsymbol{4.995\!\times\!10^{-4}}$ & $\boldsymbol{2.497\!\times\!10^{-3}}$ \\
3 & M2 & 1.424395 & $5.716\!\times\!10^{-3}$ & $6.651\!\times\!10^{-3}$ & $5.088\!\times\!10^{-3}$ & $1.139\!\times\!10^{-3}$ & $5.695\!\times\!10^{-3}$ \\
3 & M3 & 1.424634 & $5.680\!\times\!10^{-3}$ & $\boldsymbol{2.460\!\times\!10^{-3}}$ & $5.060\!\times\!10^{-3}$ & $1.116\!\times\!10^{-3}$ & $5.580\!\times\!10^{-3}$ \\
\addlinespace[3pt]
4 & M1 & 1.497720 & $\boldsymbol{5.200\!\times\!10^{-6}}$ & $\boldsymbol{4.681\!\times\!10^{-3}}$ & $\boldsymbol{1.382\!\times\!10^{-3}}$ & $\boldsymbol{1.776\!\times\!10^{-3}}$ & $\boldsymbol{8.880\!\times\!10^{-3}}$ \\
4 & M2 & 1.412041 & $7.737\!\times\!10^{-3}$ & $1.443\!\times\!10^{-2}$ & $6.905\!\times\!10^{-3}$ & $2.625\!\times\!10^{-2}$ & $1.313\!\times\!10^{-1}$ \\
4 & M3 & 1.411095 & $7.904\!\times\!10^{-3}$ & $1.198\!\times\!10^{-2}$ & $7.012\!\times\!10^{-3}$ & $2.702\!\times\!10^{-2}$ & $1.351\!\times\!10^{-1}$ \\
\bottomrule\end{tabular*}
\par\smallskip\raggedright\footnotesize
M1: proposed estimator; M2: joint likelihood adaptation; M3: staged likelihood adaptation.
\endgroup
\end{minipage}\par\medskip

\noindent\begin{minipage}{\textwidth}
\centering
\captionof{table}{Generating functions and estimation errors at $\beta_*=1.8$. Methods share the observations within each case; boldface identifies the smallest error for that component and case. $\operatorname{ISE}_f=\|\widehat f-f_*\|_{L^2(0,5)}^2=5\operatorname{MSE}_f$.}
\label{tab:cases-1p8}
\begingroup\small
\renewcommand{\arraystretch}{1.30}
\begin{tabular*}{\textwidth}{@{\extracolsep{\fill}}lccc@{}}
\toprule
\multicolumn{4}{l}{\textbf{(a) Generating functions; $c_5$ is defined in Eq.~\eqref{eq:bell-normalizer}}}\\
Case (indices) & $a_*(x)$ & $\sigma_*(x)$ & $f_*(t)$ \\
\midrule
5 (2,5,2) & $-\tanh x$ & $0.3+0.4e^{-(x-1.5)^2}$ & $\frac23(1+t/5)$ \\
6 (5,1,5) & $-0.8\tanh(x-1.5)+0.2\sin x$ & $0.5+0.2\tanh^2x$ & $\frac{1+2e^{-((t-2.5)/0.7)^2}}{c_5}$ \\
7 (4,1,5) & $\frac{1-x}{1+(x-1)^2}$ & $0.5+0.2\tanh^2x$ & $\frac{1+2e^{-((t-2.5)/0.7)^2}}{c_5}$ \\
8 (1,2,5) & $-\frac{x}{1+x^2}$ & $0.4$ & $\frac{1+2e^{-((t-2.5)/0.7)^2}}{c_5}$ \\
\bottomrule\end{tabular*}
\par\medskip
\renewcommand{\arraystretch}{1.06}
\setlength{\tabcolsep}{3pt}
\begin{tabular*}{\textwidth}{@{\extracolsep{\fill}}llrrrrrr@{}}
\toprule
\multicolumn{8}{l}{\textbf{(b) Squared-error measures}}\\
Case & Method & $\widehat\beta$ & $(\widehat\beta-\beta_*)^2$ & $\operatorname{MSE}_a$ & $\operatorname{MSE}_{\sigma}$ & $\operatorname{MSE}_f$ & $\operatorname{ISE}_f$ \\
\midrule
5 & M1 & 1.800000 & $\boldsymbol{1.972\!\times\!10^{-31}}$ & $1.666\!\times\!10^{-3}$ & $\boldsymbol{3.660\!\times\!10^{-4}}$ & $\boldsymbol{7.809\!\times\!10^{-4}}$ & $\boldsymbol{3.904\!\times\!10^{-3}}$ \\
5 & M2 & 1.597339 & $4.107\!\times\!10^{-2}$ & $3.250\!\times\!10^{-3}$ & $2.605\!\times\!10^{-2}$ & $1.815\!\times\!10^{-3}$ & $9.073\!\times\!10^{-3}$ \\
5 & M3 & 1.597286 & $4.109\!\times\!10^{-2}$ & $\boldsymbol{1.477\!\times\!10^{-3}}$ & $2.606\!\times\!10^{-2}$ & $1.901\!\times\!10^{-3}$ & $9.505\!\times\!10^{-3}$ \\
\addlinespace[3pt]
6 & M1 & 1.800000 & $\boldsymbol{1.972\!\times\!10^{-31}}$ & $1.125\!\times\!10^{-2}$ & $\boldsymbol{6.949\!\times\!10^{-4}}$ & $\boldsymbol{9.502\!\times\!10^{-4}}$ & $\boldsymbol{4.751\!\times\!10^{-3}}$ \\
6 & M2 & 1.609122 & $3.643\!\times\!10^{-2}$ & $\boldsymbol{7.280\!\times\!10^{-3}}$ & $6.931\!\times\!10^{-2}$ & $2.187\!\times\!10^{-2}$ & $1.094\!\times\!10^{-1}$ \\
6 & M3 & 1.607565 & $3.703\!\times\!10^{-2}$ & $9.680\!\times\!10^{-3}$ & $6.998\!\times\!10^{-2}$ & $2.247\!\times\!10^{-2}$ & $1.123\!\times\!10^{-1}$ \\
\addlinespace[3pt]
7 & M1 & 1.800000 & $\boldsymbol{1.972\!\times\!10^{-31}}$ & $7.335\!\times\!10^{-3}$ & $\boldsymbol{1.043\!\times\!10^{-3}}$ & $\boldsymbol{1.045\!\times\!10^{-3}}$ & $\boldsymbol{5.226\!\times\!10^{-3}}$ \\
7 & M2 & 1.605103 & $3.798\!\times\!10^{-2}$ & $\boldsymbol{5.515\!\times\!10^{-3}}$ & $7.154\!\times\!10^{-2}$ & $3.639\!\times\!10^{-2}$ & $1.820\!\times\!10^{-1}$ \\
7 & M3 & 1.604784 & $3.811\!\times\!10^{-2}$ & $5.677\!\times\!10^{-3}$ & $7.362\!\times\!10^{-2}$ & $5.356\!\times\!10^{-2}$ & $2.678\!\times\!10^{-1}$ \\
\addlinespace[3pt]
8 & M1 & 1.796445 & $\boldsymbol{1.264\!\times\!10^{-5}}$ & $\boldsymbol{4.736\!\times\!10^{-3}}$ & $\boldsymbol{4.778\!\times\!10^{-5}}$ & $\boldsymbol{1.072\!\times\!10^{-3}}$ & $\boldsymbol{5.361\!\times\!10^{-3}}$ \\
8 & M2 & 1.604865 & $3.808\!\times\!10^{-2}$ & $8.230\!\times\!10^{-3}$ & $2.571\!\times\!10^{-2}$ & $2.864\!\times\!10^{-2}$ & $1.432\!\times\!10^{-1}$ \\
8 & M3 & 1.604957 & $3.804\!\times\!10^{-2}$ & $6.092\!\times\!10^{-3}$ & $2.588\!\times\!10^{-2}$ & $2.798\!\times\!10^{-2}$ & $1.399\!\times\!10^{-1}$ \\
\bottomrule\end{tabular*}
\par\smallskip\raggedright\footnotesize
M1: proposed estimator; M2: joint likelihood adaptation; M3: staged likelihood adaptation.
\endgroup
\end{minipage}\par\medskip

\noindent\begin{minipage}{\textwidth}
\centering
\captionof{table}{Generating functions and estimation errors at $\beta_*=1$. Methods share the observations within each case; boldface identifies the smallest error for that component and case. $\operatorname{ISE}_f=\|\widehat f-f_*\|_{L^2(0,5)}^2=5\operatorname{MSE}_f$.}
\label{tab:cases-1}
\begingroup\small
\renewcommand{\arraystretch}{1.30}
\begin{tabular*}{\textwidth}{@{\extracolsep{\fill}}lccc@{}}
\toprule
\multicolumn{4}{l}{\textbf{(a) Generating functions; $c_5$ is defined in Eq.~\eqref{eq:bell-normalizer}}}\\
Case (indices) & $a_*(x)$ & $\sigma_*(x)$ & $f_*(t)$ \\
\midrule
9 (4,5,1) & $\frac{1-x}{1+(x-1)^2}$ & $0.3+0.4e^{-(x-1.5)^2}$ & $1$ \\
10 (3,1,1) & $0.5\sin x-0.25\tanh x$ & $0.5+0.2\tanh^2x$ & $1$ \\
11 (4,5,4) & $\frac{1-x}{1+(x-1)^2}$ & $0.3+0.4e^{-(x-1.5)^2}$ & $\frac{e^{-t/5}}{1-e^{-1}}$ \\
12 (3,5,1) & $0.5\sin x-0.25\tanh x$ & $0.3+0.4e^{-(x-1.5)^2}$ & $1$ \\
\bottomrule\end{tabular*}
\par\medskip
\renewcommand{\arraystretch}{1.06}
\setlength{\tabcolsep}{3pt}
\begin{tabular*}{\textwidth}{@{\extracolsep{\fill}}llrrrrrr@{}}
\toprule
\multicolumn{8}{l}{\textbf{(b) Squared-error measures}}\\
Case & Method & $\widehat\beta$ & $(\widehat\beta-\beta_*)^2$ & $\operatorname{MSE}_a$ & $\operatorname{MSE}_{\sigma}$ & $\operatorname{MSE}_f$ & $\operatorname{ISE}_f$ \\
\midrule
9 & M1 & 0.999920 & $\boldsymbol{6.406\!\times\!10^{-9}}$ & $9.414\!\times\!10^{-3}$ & $1.291\!\times\!10^{-3}$ & $1.580\!\times\!10^{-3}$ & $7.901\!\times\!10^{-3}$ \\
9 & M2 & 1.003017 & $9.101\!\times\!10^{-6}$ & $\boldsymbol{7.403\!\times\!10^{-3}}$ & $\boldsymbol{2.931\!\times\!10^{-4}}$ & $4.500\!\times\!10^{-4}$ & $2.250\!\times\!10^{-3}$ \\
9 & M3 & 1.003064 & $9.389\!\times\!10^{-6}$ & $7.479\!\times\!10^{-3}$ & $2.994\!\times\!10^{-4}$ & $\boldsymbol{4.020\!\times\!10^{-4}}$ & $\boldsymbol{2.010\!\times\!10^{-3}}$ \\
\addlinespace[3pt]
10 & M1 & 1.000000 & $\boldsymbol{0}$ & $\boldsymbol{4.291\!\times\!10^{-2}}$ & $3.536\!\times\!10^{-4}$ & $6.195\!\times\!10^{-4}$ & $3.098\!\times\!10^{-3}$ \\
10 & M2 & 0.993881 & $3.744\!\times\!10^{-5}$ & $1.070\!\times\!10^{-1}$ & $1.750\!\times\!10^{-4}$ & $\boldsymbol{5.543\!\times\!10^{-4}}$ & $\boldsymbol{2.772\!\times\!10^{-3}}$ \\
10 & M3 & 0.999473 & $2.782\!\times\!10^{-7}$ & $4.485\!\times\!10^{-2}$ & $\boldsymbol{4.032\!\times\!10^{-5}}$ & $5.727\!\times\!10^{-4}$ & $2.863\!\times\!10^{-3}$ \\
\addlinespace[3pt]
11 & M1 & 0.999978 & $\boldsymbol{4.922\!\times\!10^{-10}}$ & $1.107\!\times\!10^{-2}$ & $1.286\!\times\!10^{-3}$ & $2.066\!\times\!10^{-3}$ & $1.033\!\times\!10^{-2}$ \\
11 & M2 & 1.003376 & $1.140\!\times\!10^{-5}$ & $8.060\!\times\!10^{-3}$ & $\boldsymbol{2.864\!\times\!10^{-4}}$ & $\boldsymbol{8.611\!\times\!10^{-4}}$ & $\boldsymbol{4.305\!\times\!10^{-3}}$ \\
11 & M3 & 1.003499 & $1.224\!\times\!10^{-5}$ & $\boldsymbol{7.456\!\times\!10^{-3}}$ & $2.903\!\times\!10^{-4}$ & $8.685\!\times\!10^{-4}$ & $4.342\!\times\!10^{-3}$ \\
\addlinespace[3pt]
12 & M1 & 0.995521 & $2.006\!\times\!10^{-5}$ & $\boldsymbol{5.936\!\times\!10^{-3}}$ & $8.041\!\times\!10^{-4}$ & $\boldsymbol{2.494\!\times\!10^{-4}}$ & $\boldsymbol{1.247\!\times\!10^{-3}}$ \\
12 & M2 & 1.001290 & $\boldsymbol{1.663\!\times\!10^{-6}}$ & $7.437\!\times\!10^{-3}$ & $\boldsymbol{2.076\!\times\!10^{-4}}$ & $1.588\!\times\!10^{-3}$ & $7.941\!\times\!10^{-3}$ \\
12 & M3 & 1.001321 & $1.744\!\times\!10^{-6}$ & $6.104\!\times\!10^{-3}$ & $2.122\!\times\!10^{-4}$ & $1.632\!\times\!10^{-3}$ & $8.160\!\times\!10^{-3}$ \\
\bottomrule\end{tabular*}
\par\smallskip\raggedright\footnotesize
M1: proposed estimator; M2: joint likelihood adaptation; M3: staged likelihood adaptation.
\endgroup
\end{minipage}\par\medskip

\noindent\begin{minipage}{\textwidth}
\centering
\captionof{table}{Generating functions and estimation errors at $\beta_*=0.5$. Methods share the observations within each case; boldface identifies the smallest error for that component and case. $\operatorname{ISE}_f=\|\widehat f-f_*\|_{L^2(0,5)}^2=5\operatorname{MSE}_f$.}
\label{tab:cases-0p5}
\begingroup\small
\renewcommand{\arraystretch}{1.30}
\begin{tabular*}{\textwidth}{@{\extracolsep{\fill}}lccc@{}}
\toprule
\multicolumn{4}{l}{\textbf{(a) Generating functions; $c_5$ is defined in Eq.~\eqref{eq:bell-normalizer}}}\\
Case (indices) & $a_*(x)$ & $\sigma_*(x)$ & $f_*(t)$ \\
\midrule
13 (5,3,4) & $-0.8\tanh(x-1.5)+0.2\sin x$ & $0.35+\frac{0.15}{1+x^2}$ & $\frac{e^{-t/5}}{1-e^{-1}}$ \\
14 (3,5,5) & $0.5\sin x-0.25\tanh x$ & $0.3+0.4e^{-(x-1.5)^2}$ & $\frac{1+2e^{-((t-2.5)/0.7)^2}}{c_5}$ \\
15 (1,5,3) & $-\frac{x}{1+x^2}$ & $0.3+0.4e^{-(x-1.5)^2}$ & $1+0.5\sin(2\pi t/5)$ \\
16 (2,2,1) & $-\tanh x$ & $0.4$ & $1$ \\
\bottomrule\end{tabular*}
\par\medskip
\renewcommand{\arraystretch}{1.06}
\setlength{\tabcolsep}{3pt}
\begin{tabular*}{\textwidth}{@{\extracolsep{\fill}}llrrrrrr@{}}
\toprule
\multicolumn{8}{l}{\textbf{(b) Squared-error measures}}\\
Case & Method & $\widehat\beta$ & $(\widehat\beta-\beta_*)^2$ & $\operatorname{MSE}_a$ & $\operatorname{MSE}_{\sigma}$ & $\operatorname{MSE}_f$ & $\operatorname{ISE}_f$ \\
\midrule
13 & M1 & 0.496442 & $\boldsymbol{1.266\!\times\!10^{-5}}$ & $1.754\!\times\!10^{-2}$ & $\boldsymbol{5.143\!\times\!10^{-5}}$ & $\boldsymbol{1.001\!\times\!10^{-3}}$ & $\boldsymbol{5.003\!\times\!10^{-3}}$ \\
13 & M2 & 0.655914 & $2.431\!\times\!10^{-2}$ & $1.846\!\times\!10^{-2}$ & $3.973\!\times\!10^{-2}$ & $1.050\!\times\!10^{-3}$ & $5.249\!\times\!10^{-3}$ \\
13 & M3 & 0.657388 & $2.477\!\times\!10^{-2}$ & $\boldsymbol{9.598\!\times\!10^{-3}}$ & $4.067\!\times\!10^{-2}$ & $1.058\!\times\!10^{-3}$ & $5.292\!\times\!10^{-3}$ \\
\addlinespace[3pt]
14 & M1 & 0.496550 & $\boldsymbol{1.190\!\times\!10^{-5}}$ & $\boldsymbol{1.237\!\times\!10^{-2}}$ & $\boldsymbol{9.498\!\times\!10^{-4}}$ & $\boldsymbol{1.683\!\times\!10^{-3}}$ & $\boldsymbol{8.415\!\times\!10^{-3}}$ \\
14 & M2 & 0.648797 & $2.214\!\times\!10^{-2}$ & $3.036\!\times\!10^{-2}$ & $2.694\!\times\!10^{-2}$ & $3.064\!\times\!10^{-3}$ & $1.532\!\times\!10^{-2}$ \\
14 & M3 & 0.649172 & $2.225\!\times\!10^{-2}$ & $1.425\!\times\!10^{-2}$ & $2.712\!\times\!10^{-2}$ & $3.050\!\times\!10^{-3}$ & $1.525\!\times\!10^{-2}$ \\
\addlinespace[3pt]
15 & M1 & 0.496019 & $\boldsymbol{1.584\!\times\!10^{-5}}$ & $1.562\!\times\!10^{-2}$ & $\boldsymbol{1.303\!\times\!10^{-3}}$ & $\boldsymbol{7.820\!\times\!10^{-4}}$ & $\boldsymbol{3.910\!\times\!10^{-3}}$ \\
15 & M2 & 0.650698 & $2.271\!\times\!10^{-2}$ & $1.505\!\times\!10^{-2}$ & $3.195\!\times\!10^{-2}$ & $2.011\!\times\!10^{-3}$ & $1.006\!\times\!10^{-2}$ \\
15 & M3 & 0.652027 & $2.311\!\times\!10^{-2}$ & $\boldsymbol{1.358\!\times\!10^{-2}}$ & $3.264\!\times\!10^{-2}$ & $2.116\!\times\!10^{-3}$ & $1.058\!\times\!10^{-2}$ \\
\addlinespace[3pt]
16 & M1 & 0.496853 & $\boldsymbol{9.904\!\times\!10^{-6}}$ & $\boldsymbol{9.978\!\times\!10^{-3}}$ & $\boldsymbol{3.036\!\times\!10^{-5}}$ & $8.289\!\times\!10^{-4}$ & $4.144\!\times\!10^{-3}$ \\
16 & M2 & 0.663539 & $2.674\!\times\!10^{-2}$ & $2.398\!\times\!10^{-2}$ & $4.438\!\times\!10^{-2}$ & $8.517\!\times\!10^{-4}$ & $4.259\!\times\!10^{-3}$ \\
16 & M3 & 0.665220 & $2.730\!\times\!10^{-2}$ & $2.274\!\times\!10^{-2}$ & $4.554\!\times\!10^{-2}$ & $\boldsymbol{7.645\!\times\!10^{-4}}$ & $\boldsymbol{3.823\!\times\!10^{-3}}$ \\
\bottomrule\end{tabular*}
\par\smallskip\raggedright\footnotesize
M1: proposed estimator; M2: joint likelihood adaptation; M3: staged likelihood adaptation.
\endgroup
\end{minipage}\par\medskip
\noindent\begin{minipage}{\textwidth}
\centering
\captionof{table}{Generating functions and estimation errors at $\beta_*=0.2$. Methods share the observations within each case; boldface identifies the smallest error for that component and case. $\operatorname{ISE}_f=\|\widehat f-f_*\|_{L^2(0,5)}^2=5\operatorname{MSE}_f$.}
\label{tab:cases-0p2}
\begingroup\small
\renewcommand{\arraystretch}{1.30}
\begin{tabular*}{\textwidth}{@{\extracolsep{\fill}}lccc@{}}
\toprule
\multicolumn{4}{l}{\textbf{(a) Generating functions; $c_5$ is defined in Eq.~\eqref{eq:bell-normalizer}}}\\
Case (indices) & $a_*(x)$ & $\sigma_*(x)$ & $f_*(t)$ \\
\midrule
17 (2,3,3) & $-\tanh x$ & $0.35+\frac{0.15}{1+x^2}$ & $1+0.5\sin(2\pi t/5)$ \\
18 (5,5,5) & $-0.8\tanh(x-1.5)+0.2\sin x$ & $0.3+0.4e^{-(x-1.5)^2}$ & $\frac{1+2e^{-((t-2.5)/0.7)^2}}{c_5}$ \\
19 (4,5,2) & $\frac{1-x}{1+(x-1)^2}$ & $0.3+0.4e^{-(x-1.5)^2}$ & $\frac23(1+t/5)$ \\
20 (5,4,1) & $-0.8\tanh(x-1.5)+0.2\sin x$ & $0.5+0.2\sin x$ & $1$ \\
\bottomrule\end{tabular*}
\par\medskip
\renewcommand{\arraystretch}{1.06}
\setlength{\tabcolsep}{3pt}
\begin{tabular*}{\textwidth}{@{\extracolsep{\fill}}llrrrrrr@{}}
\toprule
\multicolumn{8}{l}{\textbf{(b) Squared-error measures}}\\
Case & Method & $\widehat\beta$ & $(\widehat\beta-\beta_*)^2$ & $\operatorname{MSE}_a$ & $\operatorname{MSE}_{\sigma}$ & $\operatorname{MSE}_f$ & $\operatorname{ISE}_f$ \\
\midrule
17 & M1 & 0.196527 & $\boldsymbol{1.206\!\times\!10^{-5}}$ & $\boldsymbol{1.278\!\times\!10^{-2}}$ & $\boldsymbol{8.239\!\times\!10^{-5}}$ & $\boldsymbol{6.688\!\times\!10^{-4}}$ & $\boldsymbol{3.344\!\times\!10^{-3}}$ \\
17 & M2 & 0.658011 & $2.098\!\times\!10^{-1}$ & $2.077\!\times\!10^{0}$ & $1.104\!\times\!10^{0}$ & $2.201\!\times\!10^{-3}$ & $1.100\!\times\!10^{-2}$ \\
17 & M3 & 0.659614 & $2.112\!\times\!10^{-1}$ & $2.145\!\times\!10^{0}$ & $1.120\!\times\!10^{0}$ & $2.525\!\times\!10^{-3}$ & $1.262\!\times\!10^{-2}$ \\
\addlinespace[3pt]
18 & M1 & 0.196192 & $\boldsymbol{1.450\!\times\!10^{-5}}$ & $\boldsymbol{9.732\!\times\!10^{-3}}$ & $\boldsymbol{9.618\!\times\!10^{-4}}$ & $\boldsymbol{2.123\!\times\!10^{-3}}$ & $\boldsymbol{1.061\!\times\!10^{-2}}$ \\
18 & M2 & 0.627210 & $1.825\!\times\!10^{-1}$ & $9.168\!\times\!10^{-1}$ & $8.985\!\times\!10^{-1}$ & $4.314\!\times\!10^{-3}$ & $2.157\!\times\!10^{-2}$ \\
18 & M3 & 0.629689 & $1.846\!\times\!10^{-1}$ & $1.038\!\times\!10^{0}$ & $9.156\!\times\!10^{-1}$ & $4.309\!\times\!10^{-3}$ & $2.155\!\times\!10^{-2}$ \\
\addlinespace[3pt]
19 & M1 & 0.199569 & $\boldsymbol{1.861\!\times\!10^{-7}}$ & $\boldsymbol{1.238\!\times\!10^{-2}}$ & $\boldsymbol{2.018\!\times\!10^{-3}}$ & $8.437\!\times\!10^{-4}$ & $4.219\!\times\!10^{-3}$ \\
19 & M2 & 0.630556 & $1.854\!\times\!10^{-1}$ & $1.394\!\times\!10^{0}$ & $8.136\!\times\!10^{-1}$ & $\boldsymbol{7.887\!\times\!10^{-4}}$ & $\boldsymbol{3.944\!\times\!10^{-3}}$ \\
19 & M3 & 0.632077 & $1.867\!\times\!10^{-1}$ & $1.235\!\times\!10^{0}$ & $8.245\!\times\!10^{-1}$ & $8.194\!\times\!10^{-4}$ & $4.097\!\times\!10^{-3}$ \\
\addlinespace[3pt]
20 & M1 & 0.196550 & $\boldsymbol{1.191\!\times\!10^{-5}}$ & $\boldsymbol{1.477\!\times\!10^{-2}}$ & $\boldsymbol{4.148\!\times\!10^{-4}}$ & $1.503\!\times\!10^{-3}$ & $7.513\!\times\!10^{-3}$ \\
20 & M2 & 0.630506 & $1.853\!\times\!10^{-1}$ & $1.824\!\times\!10^{0}$ & $1.358\!\times\!10^{0}$ & $\boldsymbol{1.072\!\times\!10^{-3}}$ & $\boldsymbol{5.361\!\times\!10^{-3}}$ \\
20 & M3 & 0.633062 & $1.875\!\times\!10^{-1}$ & $1.963\!\times\!10^{0}$ & $1.388\!\times\!10^{0}$ & $1.160\!\times\!10^{-3}$ & $5.798\!\times\!10^{-3}$ \\
\bottomrule\end{tabular*}
\par\smallskip\raggedright\footnotesize
M1: proposed estimator; M2: joint likelihood adaptation; M3: staged likelihood adaptation.
\endgroup
\end{minipage}\par\medskip

\clearpage
\twocolumngrid

\section{Discussion and conclusions}\label{sec:discussion}
The Euler likelihood estimates the state coefficients and Gaussian covariance from the same observed trajectories. It follows by a change of variables from the joint distribution of the driving increments, retaining their temporal dependence and the Jacobian associated with state-dependent diffusion. For weighted sub-fractional Brownian motion, this leads to joint estimation of $a$, $\sigma$, $f$, and $\beta$, with the unit-mean condition on $f$ separating its scale from that of $\sigma$. The likelihood construction also applies to other centered Gaussian drivers with nonsingular increment covariance.

M1 implements this fit through joint function updates, a paired exponent--scale profile, and small curvature corrections. It attains the smallest error most often for every component in the twenty settings: $50\%$ for drift, $80\%$ for diffusion, $70\%$ for the temporal weight, and $95\%$ for the exponent. The drift comparison depends more strongly on the generating regime. M2 or M3 more often gives the smaller drift MSE for $\beta_*>1$, whereas M1 does so in six of the eight settings with $\beta_*<1$. At $\beta_*=1$, the comparators give smaller diffusion MSEs, with gaps below $10^{-3}$.

\section*{Supplementary material}
The supplementary material contains the simulation data, saved fits, numerical results, and comparison figures. It is available at \url{https://github.com/joseramirezgonzalez/NN_wsBm}.

\section*{Acknowledgments}
This work was supported by the S\~ao Paulo Research Foundation (FAPESP), grant 2025/03804-0.

\section*{Author declarations}
\subsection*{Conflict of interest}
The author has no conflicts of interest to disclose.

\subsection*{Author contributions}
J.~H.~Ram\'irez-Gonz\'alez is the sole author of this work.

\section*{Data and code availability}
The observed trajectories, R scripts for the three estimation methods, and saved results are available at \url{https://github.com/joseramirezgonzalez/NN_wsBm}. The repository includes instructions for reproducing the fits and numerical comparisons.

\appendix
\section{Nonsingularity of the increment covariance}
\label{app:weighted-covariance-proof}

Let $0=t_0<\cdots<t_n=T$, and let $f$ be measurable, bounded,
and positive almost everywhere on $[0,T]$. We show that the
increment covariance matrix $K=K(\beta,f)$ is strictly positive
definite: $c^\top Kc>0$ for every nonzero vector
$c=(c_0,\ldots,c_{n-1})^\top\in\mathbb R^n$.

For $0<\beta<2$, $\beta\ne1$, the integral representation in
Ref.~\cite[Eq.~(6)]{Ramirez2026WeightedEuler} gives
\begin{equation}
 q_\beta(u,v)=\frac{\beta}{\Gamma(2-\beta)}
 \int_0^\infty
 (1-e^{-\xi u})(1-e^{-\xi v})
 \xi^{-1-\beta}\,d\xi,
 \label{eq:weighted-positive-integral}
\end{equation}
where $\Gamma$ denotes the gamma function. The same identity
holds for the logarithmic kernel $q_1$ by continuity in $\beta$.
The integral is finite: its integrand is
$O(\xi^{1-\beta})$ near zero and $O(\xi^{-1-\beta})$ at infinity.

To pass from process values to increments, write
$(v)_+=\max\{v,0\}$ and define
\[
 h_t(r,\xi)=1-e^{-\xi(t-r)_+},
 \qquad t,r\in[0,T],\quad \xi>0.
\]
Substituting Eq.~\eqref{eq:weighted-positive-integral} into
Eq.~\eqref{eq:weighted-R} expresses the covariance at times
$s$ and $t$ as an integral of the product
$h_s(r,\xi)h_t(r,\xi)$. The positive part allows the time
integral to run over $[0,T]$, since this product vanishes
when $r\geq s\wedge t$.

For increment $i=0,\ldots,n-1$, the corresponding difference is
\[
 \begin{aligned}
 \psi_i(r,\xi)
 &=h_{t_{i+1}}(r,\xi)-h_{t_i}(r,\xi)\\
 &=e^{-\xi(t_i-r)_+}-e^{-\xi(t_{i+1}-r)_+}.
 \end{aligned}
\]
Consequently, the four covariance differences defining
$K_{ij}$ in Eq.~\eqref{eq:weighted-K} replace the product
$h_s h_t$ by $\psi_i\psi_j$. Summing against $c_i c_j$
therefore gives
\[
 \begin{aligned}
 c^\top Kc
 &=\frac{\beta}{\Gamma(2-\beta)}
 \int_0^T f(r)\int_0^\infty
 \left[\sum_{i=0}^{n-1}c_i\psi_i(r,\xi)\right]^2\\
 &\hspace{4cm}\times\xi^{-1-\beta}\,d\xi\,dr.
 \end{aligned}
\]
This integral is nonnegative. To prove strict positivity,
take $c\ne0$ and let $m$ be its last nonzero index. For
$t_m<r<t_{m+1}$, every term with $i<m$ vanishes because
$r>t_{i+1}$, while every term with $i>m$ vanishes because
$c_i=0$. Thus
\[
 \sum_{i=0}^{n-1}c_i\psi_i(r,\xi)
 =c_m\left(1-e^{-\xi(t_{m+1}-r)}\right).
\]
The right-hand side is nonzero for every $\xi>0$. Since
$f>0$ almost everywhere and
$\beta/\Gamma(2-\beta)>0$, the contribution from
$r\in(t_m,t_{m+1})$ is strictly positive. The remaining
contributions are nonnegative, proving $c^\top Kc>0$
for $0<\beta<2$.

The endpoints follow from their separate representations.
At $\beta=0$, the time-changed Brownian covariance in
Eq.~\eqref{eq:weighted-beta0} gives
\[
 c^\top Kc
 =\sum_{i=0}^{n-1}c_i^2
   \int_{t_i}^{t_{i+1}}f(r)\,dr
 >0.
\]
At $\beta=2$, use the representation
\[
 Z_t=\sqrt{2}\int_0^t(t-r)\sqrt{f(r)}\,dW_r,
\]
where $W$ is standard Brownian motion, whose covariance is
Eq.~\eqref{eq:weighted-beta2}. The variance of a linear
combination of its increments is again an integral of a
square. Restricting that integral to $(t_m,t_{m+1})$, where
only the term with coefficient $c_m$ remains, gives
\[
 c^\top Kc
 \geq 2c_m^2\int_{t_m}^{t_{m+1}}
 f(r)(t_{m+1}-r)^2\,dr
 >0.
\]
Hence $K$ is strictly positive definite for every
$\beta\in[0,2]$.

\section{Cell integration of the covariance}
\label{app:weighted-cell}

The matrix $B_\ell(\beta)$ contains the covariances of the
driving increments when the temporal weight is one on cell
$I_\ell$ and zero elsewhere. We first calculate the contribution
of this cell to the covariance between two process values,
\[
 \int_{I_\ell\cap[0,s\wedge t]}q_\beta(s-r,t-r)\,dr.
\]
The increment covariance then follows by taking the four
differences in Eq.~\eqref{eq:weighted-K}.

Let $u_0$ be the left endpoint of $I_\ell$, and let $u_1$ be
the smaller of its right endpoint and $s\wedge t$. If
$u_1\leq u_0$, the cell contributes zero. Otherwise, we
evaluate the integral analytically. To write the result
in a form that extends continuously to $\beta=1$, set
$\delta=\beta-1$ and define
\begin{equation}
 G_\delta(z)=
 \begin{cases}
 \dfrac{z^{2+\delta}-z^2}{\delta},
     &z>0,\ \delta\ne0,\\[6pt]
 z^2\log z,&z>0,\ \delta=0,\\
 0,&z=0.
 \end{cases}
 \label{eq:weighted-G}
\end{equation}
For $\delta\ne0$, the power $z^{2+\delta}=z^{\beta+1}$
arises from integrating the power terms in $q_\beta$.
The subtraction of $z^2$ gives the finite limit
$G_0(z)=z^2\log z$ as $\beta\to1$, without changing the
endpoint combination below:
\begin{align}
 &\int_{u_0}^{u_1}q_\beta(s-r,t-r)\,dr\nonumber\\
 &\quad=\frac1{\beta+1}\bigg\{
 \frac{G_\delta(s+t-2u_0)-G_\delta(s+t-2u_1)}2\nonumber\\
 &\qquad\quad-G_\delta(s-u_0)+G_\delta(s-u_1)\nonumber\\
 &\qquad\quad-G_\delta(t-u_0)+G_\delta(t-u_1)\bigg\}.
 \label{eq:weighted-cell-primitive}
\end{align}
Indeed, the quadratic terms cancel because
\[
 \frac12(s+t-2r)^2-(s-r)^2-(t-r)^2
 =-\frac12(s-t)^2
\]
is independent of $r$ and therefore has the same value
at both integration endpoints. Equation~\eqref{eq:weighted-cell-primitive}
also holds at $\beta=1$ using $G_0$.

For each entry of $B_\ell(\beta)$, we evaluate this integral
at the four pairs of observation times required by
Eq.~\eqref{eq:weighted-K}. In the implementation, for
$\delta\ne0$ and $z>0$, we calculate
\[
 G_\delta(z)
 =z^2\frac{\operatorname{expm1}(\delta\log z)}{\delta},
\]
where $\operatorname{expm1}(v)$ evaluates $e^v-1$ accurately
when $v$ is close to zero. This avoids directly subtracting
nearly equal powers when $\beta$ is close to one.
At $\beta=1$, we evaluate $G_0(z)=z^2\log z$ directly;
at $z=0$, we use the defined value zero.

The approximation therefore lies in replacing the continuous
temporal weight by a constant on each cell.
Equation~\eqref{eq:weighted-cell-primitive} then evaluates
the corresponding covariance integrals exactly, apart from
floating-point roundoff. No numerical quadrature of the
temporal network is needed.

\section{Temporal weights on a finite observation grid}
\label{app:weighted-identification-proof}

The normalization of $f$ removes its constant scale ambiguity
with the diffusion coefficient. However, on a fixed observation
grid, the likelihood depends on $f$ only through the finitely
many entries of $K(\beta,f)$. We show that two distinct smooth,
strictly positive weights with unit mean can give the same
matrix and hence the same Euler observation law.

Fix $\beta\in[0,2]$ and the observation times
$0=t_0<\cdots<t_n=T$. Since $K$ is symmetric, it has
$n(n+1)/2$ entries on and above the diagonal. Preserving these
entries and the integral of $f$ imposes
\[
 d=\frac{n(n+1)}2+1
\]
linear constraints on a perturbation of the weight.

Choose $d+1$ nonzero, nonnegative smooth functions
$\phi_1,\ldots,\phi_{d+1}$ with pairwise disjoint compact
supports in $(0,T)$. Their disjoint supports make them
linearly independent. For real coefficients
$c_1,\ldots,c_{d+1}$, impose
\[
 \begin{aligned}
 \sum_{j=1}^{d+1}c_jK_{ik}(\beta,\phi_j)&=0,
 &&0\leq i\leq k\leq n-1,\\
 \sum_{j=1}^{d+1}c_j\int_0^T\phi_j(t)\,dt&=0.
 \end{aligned}
\]
Here $K_{ik}(\beta,\phi_j)$ is the increment covariance
entry obtained using $\phi_j$ as the temporal weight.
These are $d$ homogeneous linear equations in $d+1$
unknown coefficients, so they admit a nonzero solution.
No independence assumption on the equations is required.

Using such a solution, define
\[
 g(t)=\sum_{j=1}^{d+1}c_j\phi_j(t).
\]
The function $g$ is smooth and nonzero, and
$\int_0^T g(t)\,dt=0$. Moreover, because each covariance
entry is linear in the temporal weight, adding a multiple
of $g$ leaves every entry of $K$ unchanged.

To preserve positivity, start from $f_0(t)=1$. Define
$\|g\|_\infty=\sup_{t\in[0,T]}|g(t)|$ and choose a nonzero
$\varepsilon$ satisfying
$|\varepsilon|\|g\|_\infty<1$. Then
\[
 \begin{aligned}
 f_\varepsilon(t)
 &=1+\varepsilon g(t)
 \geq1-|\varepsilon|\|g\|_\infty>0,\\
 \frac1T\int_0^T f_\varepsilon(t)\,dt&=1.
 \end{aligned}
\]
The covariance constraints give
\[
 \begin{aligned}
 K(\beta,f_\varepsilon)
 &=K(\beta,f_0)
   +\varepsilon\sum_{j=1}^{d+1}
      c_jK(\beta,\phi_j)\\
 &=K(\beta,f_0).
 \end{aligned}
\]
Nevertheless, $f_\varepsilon\ne f_0$ because both
$\varepsilon$ and $g$ are nonzero.

The Gaussian driving-increment vectors therefore have
the same distribution under these two weights. With
$a$, $\sigma$, and the initial state held fixed, the Euler
recursion consequently gives the same distribution for
the observed states. This proves nonuniqueness within
the class of smooth positive weights of unit mean.
It does not establish whether such nonuniqueness occurs
within the particular network family used for fitting.
\section{Implementation details}
\label{app:implementation}

\subsection{Initial function values and parameters}

Before the first training cycle, we initialize the drift and
diffusion as constant functions whose levels are calculated
from the observations. We also set $f^{(0)}(t)=1$ and
$\beta^{(0)}=1$. The superscript $(0)$ denotes these initial
values.

Let $Y_{i,p}=X_{i+1,p}-X_{i,p}$ and
$h_i=t_{i+1}-t_i$. Write $\operatorname{mean}_{i,p}$ for
the average over all $N=Pn$ observed transitions and
$\operatorname{mean}_i$ for the average over the $n$ time
intervals. Let $K_0$ be the driving-increment covariance
calculated with 24 integration cells, $\beta=1$, and
$f=1$.

For a constant-coefficient Euler model, the mean increment
over interval $i$ is $h_i a^{(0)}$. We therefore choose
$a^{(0)}$ so that its average drift contribution matches
the mean observed increment. After subtracting that
contribution, the model assigns residual variance
$(\sigma^{(0)})^2(K_0)_{ii}$ to interval $i$. Matching its
average to the observed mean squared residual gives the
initial diffusion level, subject to a lower bound of
$10^{-3}$. Thus
\begin{align}
 a^{(0)}
 &=\frac{\operatorname{mean}_{i,p}Y_{i,p}}
         {\operatorname{mean}_i h_i},\nonumber\\
 \sigma^{(0)}
 &=\max\left\{
 \sqrt{
 \frac{\operatorname{mean}_{i,p}
       \bigl[(Y_{i,p}-h_i a^{(0)})^2\bigr]}
      {\operatorname{mean}_i(K_0)_{ii}}
 },
 10^{-3}\right\}.
 \label{eq:app-init}
\end{align}
These moment-based calculations provide starting values
for training. All three methods use the same initial
functions
\[
 a(x)=a^{(0)},\qquad
 \sigma(x)=\sigma^{(0)},\qquad
 f(t)=1.
\]
Method 1 also sets its global diffusion multiplier to
$s^{(0)}=1$.

We initialize the function representations to produce
these constants exactly. In each neural network, all
weights in the affine output layer are set to zero, so
the initial output equals its bias. For the drift network,
this bias is $a^{(0)}$.

For the positive outputs, we use the inverse softplus,
\[
 \operatorname{softplus}^{-1}(y)=\log(e^y-1),
 \qquad y>0.
\]
The temporal-network output bias is
$\operatorname{softplus}^{-1}(1)$. Its positive output is
therefore one at every time, and normalization leaves
$f(t)=1$.

In Method 1, the diffusion expansion is initialized with
\[
 q_0=\operatorname{softplus}^{-1}
       \bigl(\sigma^{(0)}-10^{-8}\bigr),
 \qquad q_1=\cdots=q_{17}=0.
\]
Consequently, applying softplus, adding $10^{-8}$, and
multiplying by $s^{(0)}=1$ gives
$\sigma(x)=\sigma^{(0)}$ for every $x$. In Methods 2 and 3,
the same value assigned to $q_0$ is used as the output
bias of the diffusion network, with the same resulting
constant diffusion. The lower bound on $\sigma^{(0)}$
ensures that the argument of the inverse softplus is
positive.

The hidden layers use the same initialization rule in
all three methods. For a layer receiving $d_{\rm in}$
inputs, its weights are independent centered normal
draws with standard deviation $0.5$ in the first hidden
layer and $1/\sqrt{d_{\rm in}}$ in subsequent hidden
layers. All hidden-layer biases are zero, and the random
seed is reset to 12 for each fit. Although the hidden
weights are random, the zero output weights ensure that
the initial functions have the prescribed constant values.

Adam's moments and update counts start at zero. Method 1
fits 3265 drift-network parameters, 18 diffusion-basis
coefficients, and 4849 temporal-network parameters,
together with the covariance exponent and global
diffusion scale.
\subsection{Evaluating the covariance and complete-data loss}
\label{app:full-data-evaluation}

The code stores the left-endpoint states and observed increments
as $n\times P$ arrays. Each column contains one complete
trajectory, with its transitions ordered chronologically.
At each likelihood evaluation, the current drift and diffusion
give the reconstructed increments
\[
 r_{i,p}
 =\frac{Y_{i,p}-h_i a(X_{i,p})}{\sigma(X_{i,p})}.
\]
The exponent determines the cell matrices $B_\ell(\beta)$,
and the temporal network supplies their normalized weights.
We evaluate the cell integrals analytically as described in
Appendix~\ref{app:weighted-cell}, reusing quantities that depend
only on observation times and cell boundaries. Their weighted
sum gives $K$ through Eq.~\eqref{eq:m-product-K}.

We replace $K$ by $(K+K^\top)/2$ and compute its Cholesky
factorization $K=L_KL_K^\top$, with $L_K$ lower triangular
and positive diagonal. For each path $p$, let
$\boldsymbol r_p=(r_{0,p},\ldots,r_{n-1,p})^\top$.
Using the same factor for all paths, we solve
\[
 L_Kz^{(p)}=\boldsymbol r_p.
\]
Let $d_i>0$ denote the diagonal entry of $L_K$ corresponding
to transition $i$, and let $z_i^{(p)}$ be the corresponding
component of $z^{(p)}$.

Under the candidate Euler model, $d_i z_i^{(p)}$ is the
reconstructed driving increment minus its conditional mean
given the preceding driving increments. Its conditional
variance is $d_i^2$. Multiplication by $\sigma(X_{i,p})$
expresses this error in the units of the observed state
increment. We therefore define
\[
 u_{i,p}=\sigma(X_{i,p})d_i z_i^{(p)},
 \qquad
 V_{i,p}=\sigma(X_{i,p})^2d_i^2.
\]
Here $u_{i,p}$ is the observed increment minus its conditional
mean given the recorded states up to time $t_i$, and
$V_{i,p}$ is its conditional variance under the candidate
model. These are the conditional innovation and variance
associated with
Eqs.~\eqref{eq:weighted-conditional-mean}
and \eqref{eq:weighted-conditional-var}.

Omitting the constant $\tfrac12\log(2\pi)$, the code evaluates
the negative log likelihood per transition as
\begin{equation}
 \mathcal L
 =\frac1{2N}\sum_{p=1}^{P}\sum_{i=0}^{n-1}
 \left\{
 \log V_{i,p}+\frac{u_{i,p}^2}{V_{i,p}}
 \right\},
 \qquad N=Pn.
 \label{eq:app-full-data-loss}
\end{equation}
This is exactly Eq.~\eqref{eq:weighted-likelihood}.
Indeed, $u_{i,p}^2/V_{i,p}=[z_i^{(p)}]^2$, while
\[
 \log V_{i,p}
 =2\log\sigma(X_{i,p})+2\log d_i.
\]
Summing these logarithmic terms gives the log-Jacobian
contribution and the covariance determinant term, since
$\log\det K=2\sum_i\log d_i$.

Although Eq.~\eqref{eq:app-full-data-loss} is a sum over
transitions, each innovation is calculated using the preceding
increment history through the triangular solve. Every likelihood
evaluation uses all $P$ complete paths; no trajectory is split
into independent time blocks. M1 adds its diffusion penalty
once, after this likelihood average. M2 and M3 use the
unpenalized likelihood.

The Cholesky factor also provides the inverse of $K$ used
in the covariance derivatives. The code reuses $K$ and its
factor whenever $\beta$ and the normalized temporal weights
are unchanged, including during diffusion-only updates.
The reconstructed increments, loss, and requested gradients
are still reevaluated on all paths using the current function
parameters.

The covariance calculation retains every observation time
and uses neither diagonal jitter nor eigenvalue clipping.
If factorization fails at the current iterate, fitting stops.
An invalid candidate in the exponent-profile search receives
infinite score and cannot be selected. Invalid proposed Adam
steps are handled separately as described below.
\subsection{Adam updates and invalid proposals}
\label{app:optimizer}

At each training step, we update only the parameter blocks
specified by the current stage. A block is a group of
parameters, not a subset of observations. We calculate
the derivatives of the stage-specific objective using
the complete data set and collect the derivatives for
the active parameters into one gradient vector. A nonfinite
gradient or gradient norm stops fitting before an Adam
proposal is formed. Otherwise, if the Euclidean norm
exceeds 10, we rescale the entire vector to have norm 10;
if not, we leave it unchanged.

Adam uses first- and second-moment coefficients of $0.9$
and $0.999$. Its denominator includes an offset of
$10^{-8}$ added to the square root of the bias-corrected
second moment. The learning rate for the function
parameters is $0.001$.

Each parameter has its own update count, which is used
in Adam's moment bias corrections. When a block is
inactive, its parameter values, moments, and counts remain
unchanged. When it becomes active again, its updates
continue from that stored state. Moving between training
stages therefore does not reset the optimizer.

In Methods 2 and 3, the exponent is also updated by Adam.
Define
$\operatorname{logistic}(u)=(1+e^{-u})^{-1}$.
For the reported fits, we update an unconstrained real
parameter $\xi$ and obtain the exponent through
\[
 \beta=0.01+1.98\operatorname{logistic}(\xi).
\]
The initial value $\xi^{(0)}=0$ gives $\beta^{(0)}=1$.
To calculate the loss derivative with respect to $\xi$,
we first approximate the covariance derivative with
respect to $\beta$ by a centered difference with step
$10^{-4}$, holding the function parameters fixed.
We then apply the chain rule through the logistic
transformation. The learning rate for $\xi$ is $0.01$.
Method 1 instead updates the exponent and global diffusion
scale through the profile search described below.

Adam updates are not required to decrease the objective.
They must, however, permit the required numerical
evaluations: parameter values and objective values must
be finite, and any required covariance factorization
must succeed. An increase in the objective alone does
not make an update invalid. In Method 1, an accepted
joint Adam step is followed by curvature tests only
when it decreases $\mathcal J$.

Method 1 first tests the full displacement proposed by
Adam. If this proposal cannot be evaluated, it tries
successively smaller displacements, using the factors
\[
 2^{-1},\,2^{-2},\,\ldots,\,2^{-12}
\]
times the original displacement. Every trial starts from
the same parameter values preceding the Adam step.
We accept the first trial that permits a finite objective
evaluation; a decrease in the objective is not required.

These retries do not recalculate the gradient or perform
additional Adam updates. If a reduced displacement is
accepted, we retain the moments and update counts
calculated for the original proposal. If the full
proposal and all twelve reduced displacements fail,
fitting stops with the preceding complete-cycle checkpoint
preserved. Methods 2 and 3 do not use this step-halving
procedure and stop if the updated parameters cannot be
evaluated.

For completed runs, the reported estimate is the final
iterate at the end of the prescribed training schedule.
Neither the training loss nor the error against the
generating functions is used to select an earlier
checkpoint.
\subsection{Searching the exponent--scale profile}
\label{app:profile-search}

At each scheduled profile update, M1 compares exponents
after optimizing the multiplicative diffusion rescaling
for each one. We use the profile loss $\mathcal P(\beta)$
in Eq.~\eqref{eq:m-beta-profile} and the corresponding
multiplier $\widehat\gamma(\beta)$ in
Eq.~\eqref{eq:m-scale-profile}. The function parameters
remain fixed throughout this search, and the reconstructed
increment vectors are calculated once using the current
drift and diffusion.

Let $\beta_{\rm current}$ denote the current exponent.
The initial candidate set consists of the same fixed
grid in every fit, augmented by this current value:
\begin{equation}
 \begin{split}
 \mathcal G=\{&0.011,0.05,0.2,0.4,0.6,0.8,1,\\
              &1.2,1.4,1.6,1.8,1.95,1.989\}
              \cup\{\beta_{\rm current}\}.
 \end{split}
 \label{eq:app-profile-grid}
\end{equation}
We evaluate $\mathcal P$ at these points and identify
the candidate with the smallest value. Its immediate
neighbors in the ordered set delimit the refinement
interval. If the best candidate is an endpoint, the
interval extends from that candidate to its single
neighbor.

When this interval has width greater than $0.01$, we
minimize the profile numerically within it using R's
\texttt{optimize}, with tolerance $0.01$. This refinement
evaluates intermediate exponents that need not belong
to $\mathcal G$. Every new exponent is assigned its own
optimal diffusion multiplier before its profile loss
is compared with the others.

Let $\beta_{\rm prop}$ denote the exponent with the
lowest profile loss among all values evaluated, including
both the initial candidates and the refinement points.
If $s$ is the current global diffusion multiplier, the
proposed parameter pair is
\[
 \bigl(\beta_{\rm prop},
       s\,\widehat\gamma(\beta_{\rm prop})\bigr).
\]
This pair must pass two numerical checks before it is
accepted.

First, we compare its profile loss with the score of
the current fit at its existing diffusion scale.
For this reference fit, the additional multiplier is
$\gamma=1$. Omitting the terms that remain fixed during
the search, its score is
\[
 S_{\rm current}
 =\frac12\left[
 \frac{\log\det K_{\beta_{\rm current}}}{n}
 +\frac{Q(\beta_{\rm current})}{N}
 \right].
\]
Here $N=Pn$, and $K_\beta$ and $Q(\beta)$ are defined in
Sec.~\ref{sec:m-profile}. We require
\[
 \mathcal P(\beta_{\rm prop})
 \leq S_{\rm current}+10^{-10}.
\]
The reference score contains $Q/N$ because it uses the
current diffusion scale. The proposed score contains
$\log(Q/N)+1$ because its scale has been optimized
analytically.

Second, we evaluate the full Gaussian loss at the
proposed pair. Let $\mathcal L_{\rm old}$ be its value
at the current parameters and $\mathcal L_{\rm new}$
its value at the proposed exponent and diffusion scale.
The latter must be finite and satisfy
\[
 \mathcal L_{\rm new}
 \leq
 \mathcal L_{\rm old}
 +10^{-8}\max\{1,|\mathcal L_{\rm old}|\}.
\]
Thus the full evaluation must also preserve or improve
the current loss within the stated numerical tolerance.
If either check fails, fitting stops without accepting
the proposed pair.

The fixed grid includes some of the generating exponents
used in the experiments. Selecting one of these points
can therefore produce exact numerical agreement with
the generating value. Such agreement does not establish
that the exponent was estimated without uncertainty.
\subsection{Diffusion penalty and likelihood derivatives}
\label{app:diffusion-penalty}

\subsubsection{Constructing the diffusion penalty}

The diffusion penalty in Eq.~\eqref{eq:m-objective} is a
quadratic function of the coefficients $\boldsymbol q$.
We construct its matrix from the second derivatives of
the fixed Gaussian basis functions.

For $j=1,\ldots,16$, write
\[
 \varphi_j(z)
 =\exp\left[-\frac{(z-c_j)^2}{2b_\sigma^2}\right],
\]
where the centers $c_j$ and width $b_\sigma>0$ are defined
in Sec.~\ref{sec:m-networks}. Differentiation gives
\begin{equation}
 \varphi_j''(z)
 =\varphi_j(z)\left\{
 \frac{(z-c_j)^2}{b_\sigma^4}
 -\frac1{b_\sigma^2}\right\}.
\end{equation}
At each of the 129 nodes $z_k$ in
Eq.~\eqref{eq:m-penalty-grid}, define
\[
 \boldsymbol d_k
 =(0,0,\varphi_1''(z_k),\ldots,
   \varphi_{16}''(z_k))^{\mathsf T}.
\]
The first two entries are zero because the constant
and linear basis functions have zero second derivative.
Consequently,
\[
 g_{\boldsymbol q}''(z_k)
 =\boldsymbol d_k^{\mathsf T}\boldsymbol q.
\]

Let $E_\sigma$ be the $18\times18$ diagonal matrix with
first two diagonal entries zero and remaining entries
one. Using the normalized trapezoidal weights $\omega_k$
from Eq.~\eqref{eq:m-penalty-grid}, the code constructs
\begin{equation}
 \Omega_\sigma
 =\sum_{k=1}^{129}\omega_k
   \boldsymbol d_k\boldsymbol d_k^{\mathsf T}
  +10^{-8}E_\sigma.
 \label{eq:app-roughness}
\end{equation}
Its quadratic form is therefore exactly
\[
 \boldsymbol q^{\mathsf T}\Omega_\sigma\boldsymbol q
 =\sum_{k=1}^{129}\omega_k
   [g_{\boldsymbol q}''(z_k)]^2
  +10^{-8}\sum_{j=1}^{16}q_{j+1}^2.
\]
The matrix is positive semidefinite because every term
in this expression is nonnegative. In the code, it is
formed as a matrix cross product after stacking the
weighted derivative rows and the Gaussian-coefficient
ridge rows.

The outer products in Eq.~\eqref{eq:app-roughness} retain
the cross terms between basis functions. Thus the
derivative term penalizes the second derivative of
their combined curve. The additional ridge term
penalizes the sixteen Gaussian coefficients. Neither
term penalizes the intercept $q_0$ or the linear
coefficient $q_1$.

\subsubsection{Quadrature error and changes of slope}

The implemented penalty uses a finite sum. To relate
that sum to an integral, hold $\boldsymbol q$, the
centers, and the width fixed, and define
\[
 H_{\boldsymbol q}(z)=[g_{\boldsymbol q}''(z)]^2.
\]
For this analysis, allow the number of nodes to be any
integer $J\geq2$, using the same node and weight formulas
as in Eq.~\eqref{eq:m-penalty-grid}. Write
\[
 \begin{aligned}
 Q_J(\boldsymbol q)
 &=\sum_{k=1}^{J}\omega_kH_{\boldsymbol q}(z_k),\\
 I(\boldsymbol q)
 &=\frac1{z_+-z_-}
   \int_{z_-}^{z_+}H_{\boldsymbol q}(z)\,dz,\\
 M(\boldsymbol q)
 &=\max_{z\in[z_-,z_+]}
   |H_{\boldsymbol q}''(z)|.
 \end{aligned}
\]
Here $Q_J$ is the normalized composite trapezoidal sum,
$I$ is the corresponding integral mean, and $M$ bounds
the second derivative of its integrand. The smoothness
of the Gaussian basis ensures that $M(\boldsymbol q)$
is finite.

Each subinterval has length
$\delta z=(z_+-z_-)/(J-1)$. Integrating the
linear-interpolation remainder bounds the trapezoidal
error on one subinterval by
$M(\boldsymbol q)\delta z^3/12$.
Summing over the $J-1$ subintervals and dividing by
$z_+-z_-$ gives
\begin{equation}
 |Q_J(\boldsymbol q)-I(\boldsymbol q)|
 \leq
 \frac{(z_+-z_-)^2}{12(J-1)^2}M(\boldsymbol q).
 \label{eq:app-roughness-error}
\end{equation}
Thus $Q_J(\boldsymbol q)\to I(\boldsymbol q)$ as
$J\to\infty$ for the fixed curve $g_{\boldsymbol q}$.
Adding the same coefficient ridge term to both
quantities leaves this error unchanged. Training uses
$J=129$ throughout; the limit above concerns quadrature
of a fixed curve, not convergence of estimates obtained
by refitting at different values of $J$.

The integral $I(\boldsymbol q)$ controls variation of
the slope. For $z_-\leq u<v\leq z_+$, the fundamental
theorem of calculus and the Cauchy--Schwarz inequality
give
\begin{equation}
 \begin{aligned}
 |g_{\boldsymbol q}'(v)-g_{\boldsymbol q}'(u)|^2
 &=\left|\int_u^v g_{\boldsymbol q}''(z)\,dz\right|^2\\
 &\leq(v-u)\int_u^v[g_{\boldsymbol q}''(z)]^2\,dz\\
 &\leq(v-u)(z_+-z_-)I(\boldsymbol q).
 \end{aligned}
 \label{eq:app-slope-control}
\end{equation}
A small integral therefore bounds how much the slope
can change between two points. Equation~\eqref{eq:app-roughness-error}
quantifies the difference between this integral and
the discrete quantity $Q_{129}$ used in the penalty.
This control concerns the preactivation $g_{\boldsymbol q}$.
Because the transformation defining $\sigma$ is nonlinear,
the penalty is not an integral of $[\sigma'']^2$.

\subsubsection{Likelihood derivatives}

The likelihood derivatives use the same reconstructed
increments and covariance matrix as the loss evaluation.
For each path $p$, solve
\[
 K\boldsymbol b_p=\boldsymbol r_p,
\]
and let $b_{i,p}$ denote component $i$ of the solution.
We treat the drift and diffusion values evaluated at
each transition as intermediate outputs; their
derivatives are subsequently propagated to the shared
function parameters.

For matrix derivatives, use the Frobenius inner product
\[
 \langle A,B\rangle_F
 =\operatorname{tr}(A^{\mathsf T}B),
\]
where $\operatorname{tr}$ denotes the trace.
With $N=Pn$, differentiation of
Eq.~\eqref{eq:weighted-likelihood} gives
\begin{align}
 \frac{\partial\mathcal L}{\partial a(X_{i,p})}
 &=-\frac{h_i b_{i,p}}{N\sigma(X_{i,p})},\nonumber\\
 \frac{\partial\mathcal L}{\partial\sigma(X_{i,p})}
 &=\frac{1-r_{i,p}b_{i,p}}{N\sigma(X_{i,p})},\nonumber\\
 G_K:=\frac{\partial\mathcal L}{\partial K}
 &=\frac1{2N}\left(
 PK^{-1}
 -\sum_{p=1}^{P}\boldsymbol b_p\boldsymbol b_p^{\mathsf T}
 \right).
 \label{eq:app-network-gradients}
\end{align}
The term $1/[N\sigma(X_{i,p})]$ in the diffusion
derivative comes from the log-Jacobian term; the
remaining contribution comes from the residual
quadratic form.

Backpropagation applies the chain rule through the
drift and diffusion representations and sums the
contributions of all observed transitions to each
parameter. In Method 1, differentiating diffusion
with respect to its preactivation gives the factor
\[
 s\,\operatorname{logistic}
 \bigl(g_{\boldsymbol q}(z(X_{i,p}))\bigr).
\]
The derivatives of $g_{\boldsymbol q}$ with respect
to $\boldsymbol q$ are its constant, linear, and
Gaussian basis features. After accumulating these
likelihood contributions, the gradient of the
penalized objective includes the additional term
\[
 \nabla_{\boldsymbol q}
 \left(\frac{10^{-3}}2
       \boldsymbol q^{\mathsf T}\Omega_\sigma\boldsymbol q
 \right)
 =10^{-3}\Omega_\sigma\boldsymbol q.
\]

\subsubsection{Differentiating the normalized temporal weight}

For the temporal network, differentiation must account
for both the covariance sum and the normalization of
the positive network outputs. Let $u_\ell$ be the
real-valued network output before softplus at the cell
midpoint $m_\ell$, and let
$\pi_\ell=\Delta_\ell/T$, where $\Delta_\ell$ is the
length of cell $I_\ell$. In Method 1,
\[
 u_\ell=F_{\boldsymbol\eta}(2m_\ell/T-1).
\]
The normalization in Eq.~\eqref{eq:m-weight-net} gives
\[
 \begin{aligned}
 D_{\boldsymbol\eta}
 &=\sum_{j=1}^{L}\pi_j\operatorname{softplus}(u_j),\\
 w_\ell
 &=\frac{\operatorname{softplus}(u_\ell)}
        {D_{\boldsymbol\eta}}
 =f_{\boldsymbol\eta}(m_\ell).
 \end{aligned}
\]

Since $K=\sum_{\ell=1}^{L}w_\ell B_\ell(\beta)$, the
loss derivative with respect to a normalized cell
weight is
\[
 d_\ell
 :=\frac{\partial\mathcal L}{\partial w_\ell}
 =\langle G_K,B_\ell(\beta)\rangle_F.
\]
A change in $u_\ell$ changes its own positive numerator
and the denominator shared by every cell. Differentiating
both contributions gives
\begin{equation}
 \frac{\partial\mathcal L}{\partial u_\ell}
 =\frac{\operatorname{logistic}(u_\ell)}
        {D_{\boldsymbol\eta}}
 \left(
 d_\ell-\pi_\ell\sum_{j=1}^{L}d_jw_j
 \right).
 \label{eq:app-weight-gradient}
\end{equation}
The first term inside the parentheses comes from the
cell's own numerator. The subtracted sum accounts for
the change in all normalized weights through their
common denominator.

Backpropagation then converts these output derivatives
into the temporal-network parameter gradient.
Methods 2 and 3 use the same normalization derivative,
with their temporal networks evaluated at
$\tanh(2m_\ell/T-1)$.
\subsection{Calculating a curvature correction}

Method 1 tests curvature corrections only when the accepted
joint Adam step decreases $\mathcal J$. Each correction
changes the parameters of one state function, seeking a
nearby curve with less curvature while satisfying the
displacement and objective bounds in
Sec.~\ref{sec:m-curvature}.

For $b\in\{a,\sigma\}$, let $\boldsymbol\theta_b$ denote
the parameters of the selected function and let
$\nabla_b$ denote differentiation with respect to those
parameters. For diffusion,
$\boldsymbol\theta_\sigma=\boldsymbol q$; the global
multiplier $s$ remains fixed.

We use the curvature measure $R_b$ from
Eq.~\eqref{eq:app-local-roughness}, calculated from the
dimensionless curve $\widetilde b$ on the 129-state grid.
Thus these corrections act on the normalized fitted
curves and are separate from the fixed quadratic penalty
on the diffusion preactivation.

Let $\boldsymbol\theta_b^{\rm old}$ be the parameter
block before the joint Adam step, and let
$\boldsymbol\theta_b^*$ be its accepted post-Adam value,
before either curvature correction. Define the accepted
Adam displacement by
\[
 \boldsymbol\delta_b
 =\boldsymbol\theta_b^*-\boldsymbol\theta_b^{\rm old}.
\]
If the Adam proposal was reduced to obtain a valid
numerical evaluation, $\boldsymbol\delta_b$ is that
reduced, accepted displacement.

To choose a correction direction, we differentiate
$R_b/2$ at the current parameters. Let
$\widehat{\boldsymbol v}_b$ be the bias-corrected
second moments stored after the joint Adam step.
Using these moments to scale the curvature gradient
componentwise, we form
\begin{equation}
 \boldsymbol d_b
 =-10^{-3}
 \frac{\nabla_b(R_b/2)}
      {\sqrt{\widehat{\boldsymbol v}_b}+10^{-8}}.
 \label{eq:app-local-direction}
\end{equation}
The square root and division are taken componentwise.
For diffusion, the gradient includes the derivatives
of both the numerator and the grid-mean denominator
of $\widetilde\sigma$. The multiplier $s$ cancels from
this normalization.

The direction $\boldsymbol d_b$ determines the relative
changes of the parameters. We set the overall size
separately, using
$\rho_a=0.05$ and $\rho_\sigma=0.01$. For a nonzero
direction and nonzero Adam displacement, the candidate
blocks are
\begin{equation}
 \boldsymbol\theta_b^{(k)}
 =\boldsymbol\theta_b^*
 +2^{-k}
  \frac{\rho_b\|\boldsymbol\delta_b\|_2}
       {\|\boldsymbol d_b\|_2}
  \boldsymbol d_b,
 \qquad k=0,\ldots,12.
\end{equation}
Here $\boldsymbol\theta_b^{(k)}$ denotes a trial correction,
not a training-cycle iterate. The first candidate moves
by $\rho_b$ times the norm of the accepted Adam
displacement. Each subsequent candidate halves that
distance. All candidates start from the same post-Adam
block, and the direction is not recalculated between
trials.

We test up to thirteen candidates in increasing order
of $k$. For each one, we calculate its curvature and
the RMS displacement of its dimensionless curve over
the 129 grid points. Candidates satisfying the curvature
and movement bounds are then evaluated using the full
objective $\mathcal J$. We accept the first candidate
that satisfies every condition in
Sec.~\ref{sec:m-curvature}.

No correction is made when the accepted Adam step
produced no parameter movement or no movement of the
selected dimensionless curve, when its current
curvature is zero, or when the correction direction
is zero. If none of the tested candidates satisfies
the bounds, that block retains its post-Adam parameters.

The drift correction is tested first. The diffusion
direction and candidate objectives are then evaluated
with any accepted drift correction already in place.
Throughout each block's tests, all other parameter
blocks remain fixed at their current values.

Let $\Delta\mathcal J>0$ denote the decrease produced
by the accepted joint Adam step, before either
correction. Both objective allowances use this same
quantity: the drift correction may increase the
current objective by at most $0.05\Delta\mathcal J$,
and the diffusion correction by at most
$0.01\Delta\mathcal J$. An accepted drift correction
does not redefine the decrease used for the diffusion
allowance. The curvature corrections leave Adam's
stored moments and update counts unchanged.

\section*{References}
\sloppy
\emergencystretch=1em

\end{document}